\documentclass[11pt]{article}

\usepackage[final]{acl}

\usepackage{times}
\usepackage{latexsym}

\usepackage[T1]{fontenc}

\usepackage[utf8]{inputenc}

\usepackage{microtype}

\usepackage{inconsolata}

\usepackage{graphicx}

\usepackage{amssymb}
\usepackage{amsmath}
\usepackage{subcaption}
\usepackage{booktabs}
\usepackage{multirow}
\usepackage{adjustbox}
\usepackage{enumitem}
\usepackage[table]{xcolor}
\usepackage{wrapfig}
\usepackage{float}

\definecolor{oursblue}{RGB}{232,240,252}

\title{Rethinking Parameter Sharing for LLM Fine-Tuning with Multiple LoRAs}

\author{Hao Ban \and Kaiyi Ji\thanks{Corresponding Author} \\ 
        Department of Computer Science and Engineering, University at Buffalo \\
        \{haoban, kaiyiji\}@buffalo.edu}

\begin{document}
\maketitle
\begin{abstract}
Large language models are often adapted using parameter-efficient techniques such as Low-Rank Adaptation (LoRA), formulated as $y = W_0x + BAx$, where $W_0$ is the pre-trained parameters and $x$ is the input to the adapted layer.
 While multi-adapter extensions often employ multiple LoRAs, prior studies suggest that the inner $A$ matrices are highly similar during training and thus suitable for sharing. We revisit this phenomenon and find that this similarity is largely attributable to the identical initialization rather than shared knowledge, with $B$ playing a more critical role in knowledge encoding and transfer. Motivated by these insights, we propose \textbf{ALoRA}, an asymmetric multi-LoRA design with multiple $A$ matrices and a single shared $B$ in multi-task fine-tuning, and \textbf{Fed-ALoRA}, which shares $B$ across clients in federated fine-tuning under both homogeneous and heterogeneous settings, through a novel matrix decomposition strategy to accommodate heterogeneous ranks across clients. Experiments on commonsense reasoning, math reasoning, multi-task NLP dataset, and federated NLP dataset demonstrate that our methods achieve more balanced performance across tasks with comparable or superior average accuracy relative to existing multi-LoRA approaches. The code is available at \url{https://github.com/OptMN-Lab/ALoRA}.
\end{abstract}

\section{Introduction}

Large language models (LLMs) have achieved remarkable performance across diverse domains \citep{comanici2025gemini,dubey2024llama,achiam2023gpt}, but the growing scale makes conventional full fine-tuning increasingly expensive. Parameter-efficient fine-tuning (PEFT) addresses this challenge by freezing the pre-trained model and updating only a small subset of parameters, improving efficiency while maintaining performance \citep{han2024parameterefficient}. Among PEFT methods, Low-Rank Adaptation (LoRA) \citep{hulora} is particularly popular: it decomposes weight updates into trainable low-rank matrices $A$ and $B$, which can be merged into the pre-trained model without extra inference latency. 

Recent studies have shown that a single LoRA has limited capacity when handling diverse data distributions \citep{cai2025survey,yang2024low}. A natural extension is to use multiple LoRAs, where each module can specialize in different data modes such as tasks, domains, and distributed clients \citep{liao2025hmora,sunstronger,li2024mixlora,wumixture}. In multi-task fine-tuning, adapters are required to handle task heterogeneity \citep{liang2025thanora}, and in federated fine-tuning, they should account for client heterogeneity and personalization \citep{bian2025fedalt}. However, naively employing multiple LoRAs also increases computation and communication costs, which makes this approach less efficient.

To address this problem, recent methods explore parameter sharing across LoRA modules to improve parameter efficiency. HydraLoRA \citep{tian2024hydralora} observes that $A$ matrices trained on different tasks exhibit very high similarity, and proposes a single shared $A$ with multiple $B$s for multi-task fine-tuning. FedSA-LoRA \citep{guoselective} reports similar findings in federated fine-tuning and transmits only $A$ matrices for server aggregation with reduced communication costs.  These studies attribute the high similarity in $A$ matrices to the shared knowledge. 

In this paper, we revisit this similarity phenomenon and find that the similarity of $A$ stems mainly from identical initialization rather than shared knowledge. Our analysis of the learning dynamics suggests that $A$ tends to act as a feature projector, whereas $B$ captures domain specific knowledge. Empirically, we observe that sharing $B$ often leads to more effective knowledge transfer than sharing $A$ in both multi-task and federated fine-tuning settings. These insights motivate an interesting but underexplored question: {\em might sharing the module $B$, rather than $A$, be more effective for parameter and knowledge sharing?} In this paper, we provide a positive answer, with our main contributions as follows.
\begin{list}{$\bullet$}{\topsep=0.3ex \leftmargin=0.1in \rightmargin=0.in \itemsep =0.022in} 
\item We propose \textbf{ALoRA}, an asymmetric multi-LoRA architecture for multi-task fine-tuning. It employs multiple $A$ matrices and a single shared $B$ matrix, where the $A$ matrices are dynamically routed by the inputs. This design enables each $A$ to explore distinct feature subspaces while encouraging knowledge transfer through the shared $B$.
    \item We propose \textbf{Fed-ALoRA}, which communicates only $B$ matrices rather than full LoRA parameters for aggregation on server. It supports both homogeneous and heterogeneous settings with the same and different ranks across clients, whereas existing parameter-sharing federated fine-tuning methods focus only on the homogeneous case. In the homogeneous setting, Fed-ALoRA updates all $A$ matrices locally, and transmits and aggregates only $B$ matrices on server side. In the heterogeneous setting, direct aggregation of $B$ is infeasible due to their distinct sizes, so we decompose $B$ into $(B_1,B_2)$ with appropriate sizes and introduce an auxiliary matrix for further dimension adjustment. Compared to full LoRA aggregation, Fed-ALoRA reduces communicated parameters by up to 50\% and 75\%  in the homogeneous and heterogeneous settings, respectively, while maintaining performance. 

    \item We conduct extensive experiments on intra-domain multi-task benchmarks such as commonsense and math reasoning, cross-domain multi-task NLP dataset, and federated NLP dataset, using major open-source models such as LLaMA and Qwen.
    Across all datasets, our methods consistently deliver more balanced performance with comparable or superior accuracy compared to existing methods. In particular, \textbf{ALoRA} surpasses the sharing-$A$ approach \textbf{HydraLoRA} on LLaMA2-7B, improving average ROUGE-1 by +0.68 with a $\Delta m\%$ (which quantifies performance balance via mean drop from single-objective baselines) gain of -1.94. Similarly, \textbf{Fed-ALoRA} outperforms the sharing-$A$ approach \textbf{FedSA-LoRA}, achieving gains of +1.26 (homogeneous) and +1.96 (heterogeneous) with $\Delta m\%$ gains of -2.08 and -2.65, respectively. Similar improvements can also be observed on Qwen2-7B. Compared with approaches that aggregate full LoRA parameters, our method attains comparable performance, smaller $\Delta m\%$, and substantially reduced communication cost by transmitting much fewer parameters.
    
\end{list}

\vspace{-0.1cm}

\section{Background}
\vspace{-0.1cm}
\subsection{Low-Rank Adaptation}
Pre-trained language models exhibit low intrinsic dimensionality when adapted to downstream tasks \citep{aghajanyan2021intrinsic}. LoRA leverages this property by approximating weight updates through low-rank decomposition. Particularly, for a pre-trained weight matrix $W_0\in\mathbb{R}^{d_{\text{out}}\times d_{\text{in}}}$, the weight updates is defined as $\Delta W=BA$, where $A\in\mathbb{R}^{r\times d_{\text{in}}}$, $B\in\mathbb{R}^{d_{\text{out}}\times r}$, and the rank $r\ll min(d_{\text{in}},d_{\text{out}})$. During training, only $A$ and $B$ matrices are trainable. 
In practice, $A$ is typically initialized using Kaiming Uniform \citep{he2015delving}, and $B$ is initialized as zero. 

\vspace{-0.1cm}
\subsection{Fine-Tuning with Multiple LoRAs}\label{sec:ftmultilora}

Multiple LoRA-based methods extend vanilla LoRA with additional modules to improve adaptability across heterogeneous domains~\citep{dettmers2023qlora,zi2023delta}. In multi-task fine-tuning, they often use MoE designs where LoRAs act as dynamically routed experts~\citep{huanglorahub,luo2024moelora}, while in federated fine-tuning, they aim to balance personalization with shared knowledge aggregation~\citep{raje2025ravan,zhang2025fed}. A common idea among these multi-LoRA approaches is to share the same matrix $A$, based on the observation that $A$ matrices from LoRAs trained on different tasks or clients are often highly similar. For example, HydraLoRA \citep{tian2024hydralora} employs a single $A$ matrix and multiple $B$ matrices to express the weight updates: 
    $\Delta W=\sum_{i=1}^n w_iB_iA,$ 
where $n$ is the number of $B$ matrices, $w_i$ is the gating score for each $B_i$. The federated multi-LoRA approach 
FedSA-LoRA \citep{guoselective} shares only the $A$ matrices for server aggregation, after which the server broadcasts the aggregated $A$ to all clients. The model update of client $i$ is given by: $\Delta W_i^t = B_i^t \bar{A^t}$, and $\bar{A^t}=\text{Agg}(A_1^{t-1},\cdots,A_n^{t-1})$, where $n$ is the number of clients, and $t$ is the current communication round, and $\text{Agg}(\cdot)$ denotes an aggregation algorithm such as simple averaging.

\vspace{-0.1cm}
\section{Revisiting Parameter Sharing in Multi-LoRA Fine-tuning}\label{sec:revisiting}
\vspace{-0.1cm}

As noted earlier, a common strategy for parameter sharing is to reuse the same matrix $A$ across multiple LoRA modules, with the goal of reducing the total number of parameters and enabling knowledge transfer. In this section, we systematically re-examine this approach through a series of controlled experiments. Full implementation details and similar results on Qwen2 model are provided in Appendix~\ref{appx:multilora}.

\subsection{Similarity in $A$ Stems from Same Initialization, Not Shared Knowledge}\label{subsec:similarity}  
A primary motivation for sharing $A$ across LoRA modules is the observation that the matrices $A_i$ of different LoRAs often appear similar during training. However, upon closer examination, we find that this similarity largely arises from their common initialization rather than from the shared knowledge.  
We fine-tune the LLaMA2-7B model \citep{touvron2023llama} separately on classification and summarization tasks from the Dolly-15K dataset \citep{DatabricksBlog2023DollyV2}, using either identical or different random seeds for $A$ initialization (leading to different initializations for the matrices $A_i$), and compare the resulting LoRA modules using principal angle-based similarity \citep{zhu2013angles}, where a value of $1$ indicates complete similarity and $0$ indicates dissimilarity. The results are shown in Figure~\ref{fig:fig1a}-\ref{fig:fig1c}, and details of the similarity metric are discussed in Appendix~\ref{appx:similarity}.

\vspace{0.1cm}
\noindent{\bf Observation.} Figure~\ref{fig:fig1a} shows that with the same initialization, $A$ matrices are highly similar. In contrast, Figure~\ref{fig:fig1b}-\ref{fig:fig1c} show that with different initializations, $A$ matrices from either the same or different tasks exhibit little similarity, while $B$ matrices display relatively higher similarity. These results suggest that the $A$ is highly sensitive to random seeds rather than necessarily capturing shared knowledge, whereas $B$ is less affected. 

\begin{figure}[t]
\centering
\begin{subfigure}[t]{0.49\linewidth}
  \centering
  \includegraphics[width=\linewidth]{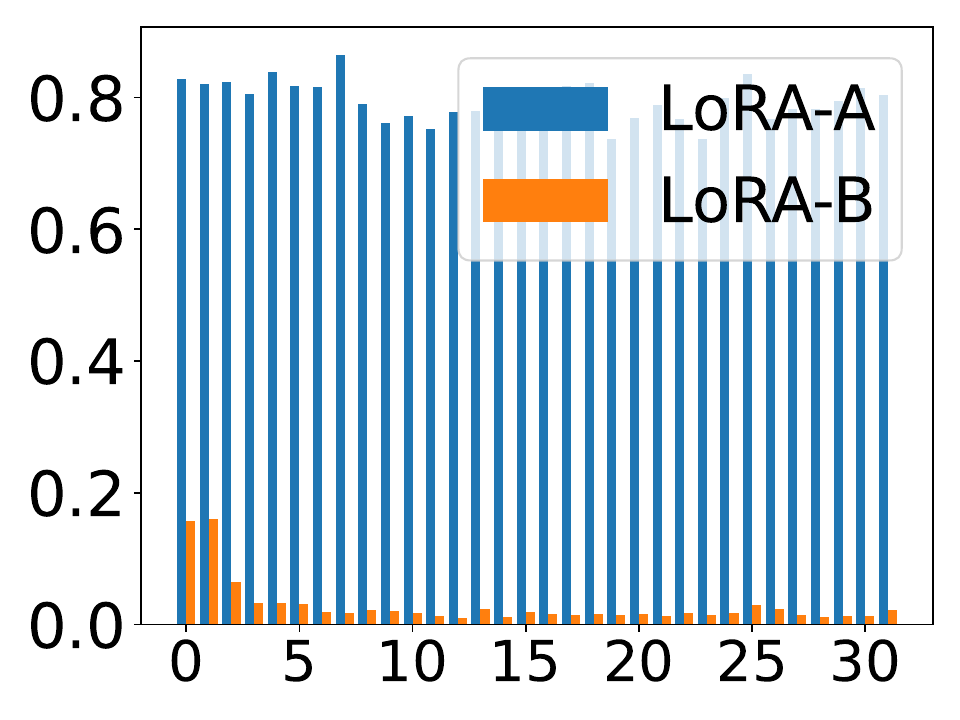}
  \caption{Different tasks, same init.}
  \label{fig:fig1a}
\end{subfigure}\hfill
\begin{subfigure}[t]{0.49\linewidth}
  \centering
  \includegraphics[width=\linewidth]{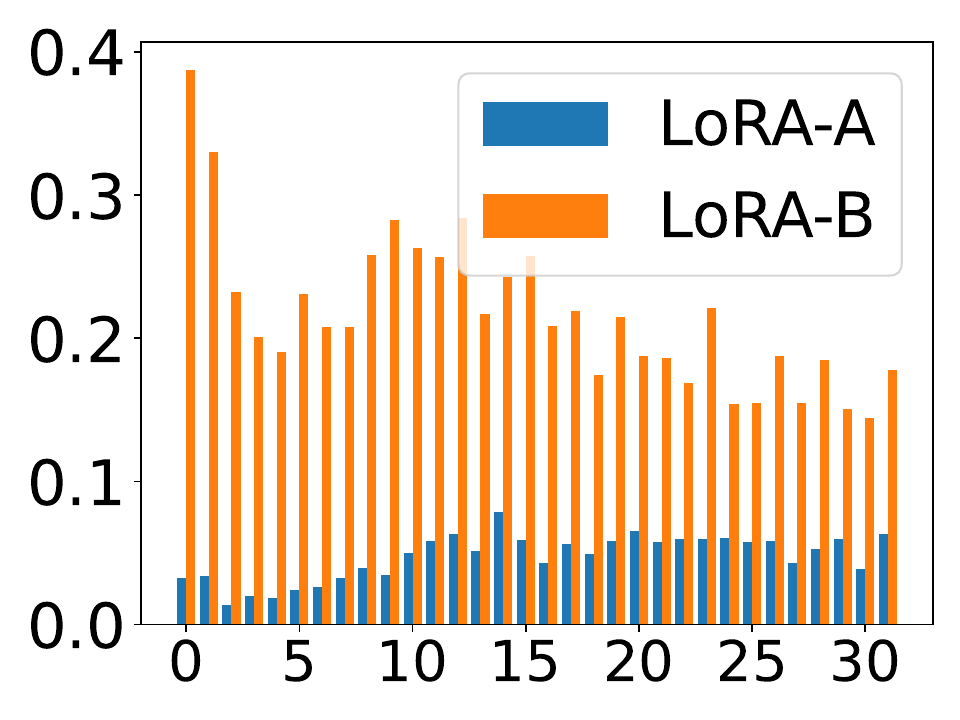}
  \caption{Same task, different init.}
  \label{fig:fig1b}
\end{subfigure}
\begin{subfigure}[t]{0.49\linewidth}
  \centering
  \includegraphics[width=\linewidth]{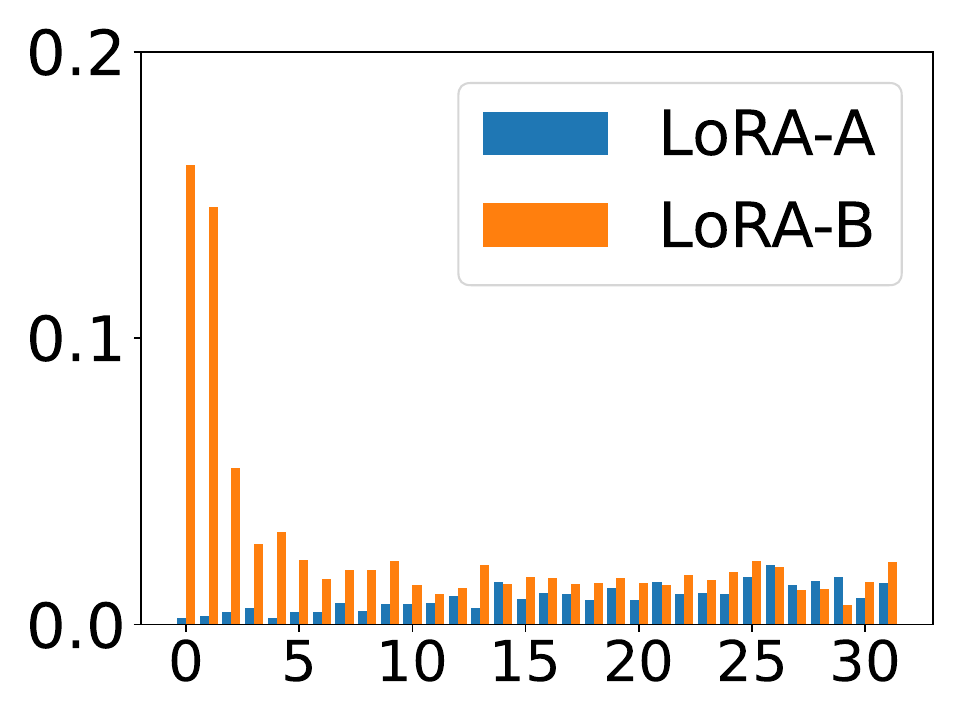}
  \caption{Different tasks and init.}
  \label{fig:fig1c}
\end{subfigure}\hfill
\begin{subfigure}[t]{0.49\linewidth}
  \centering
  \includegraphics[width=\linewidth]{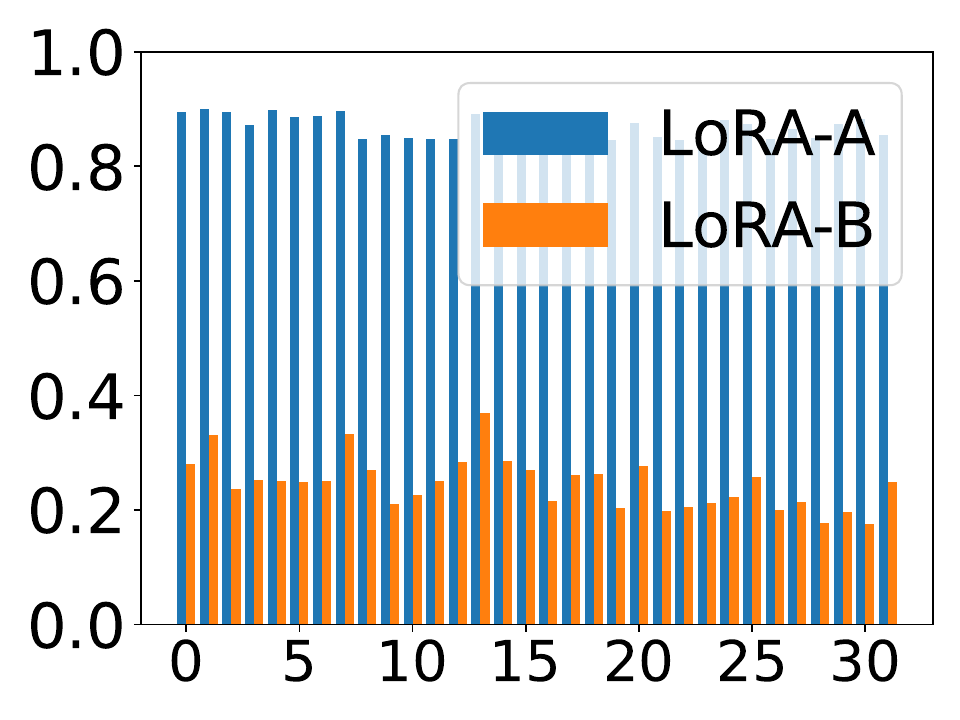}
  \caption{Before and after FT.}
  \label{fig:fig1d}
\end{subfigure}
\vspace{-0.1cm}
\caption{Layer-wise similarity analysis. The x-axis denotes the layer indices, and the y-axis denotes the similarity scores. Subfigures (a)-(c) compare different LoRA modules under varying tasks and initialization settings: (a) two different tasks with the same random seed; (b) the same task with different random seeds; and(c) two different tasks with different random seeds. Subfigure (d) compares the same LoRA module before and after fine-tuning. $A$ matrices are similar only under the same initialization, whereas $B$ exhibits relatively stable similarity across different tasks and seeds. In addition, $A$ remains largely unchanged from initialization.}
\label{fig:fig1}
\vspace{-0.2cm}
\end{figure}

\subsection{Dissecting Distinct Dynamics of $A$ and $B$ During Training}\label{subsec:dynamics}

The above analysis motivates us to further investigate the learning dynamics of $A$ and $B$ during training by comparing their states before and after fine-tuning on the summarization task\footnote{We use the checkpoints from the second and final steps, since $B$ is initialized to zero at the beginning.}.
Our experiments evaluate (i) the similarity of modules $A$ and $B$ (using the similarity metric in Section~\ref{subsec:similarity}) and (ii) the magnitude and directional variations of $\Delta W$, $A$ and $B$. To formalize this\footnote{We follow the same setup as in \citet{liu2024dora}.}, any weight matrix $W\in\mathbb{R}^{d_{\text{out}}\times d_{\text{in}}}$ can be decomposed into a magnitude and a direction component: 
$W = \|W\|_c \frac{W}{\|W\|_c} = m V,$
where $\|\cdot\|_c$ denotes the column-wise norm. Here, $m\in\mathbb{R}^{1\times d_{\text{in}}}$ is the magnitude vector, with $m_j$ denoting the norm of the $j$-th column of $W$, and $V\in\mathbb{R}^{d_{\text{out}}\times d_{\text{in}}}$ is the direction matrix with unit-norm columns.   
Given two matrices $W_1$ and $W_2$, their magnitude and direction discrepancies are defined as 

\vspace{-0.4cm}

{\small\begin{align*}
\Delta M =& \frac{1}{d_\text{in}} \sum_{j=1}^{d_\text{in}} |m_{1,j}-m_{2,j}|
\\\Delta D =&\frac{1}{d_\text{in}} \sum_{j=1}^{d_\text{in}} (1-\cos(V_{1,j}, V_{2,j})).
\end{align*}}

\vspace{-0.1cm}


\begin{figure}[t]
\begin{center}
\includegraphics[width=0.49\linewidth]{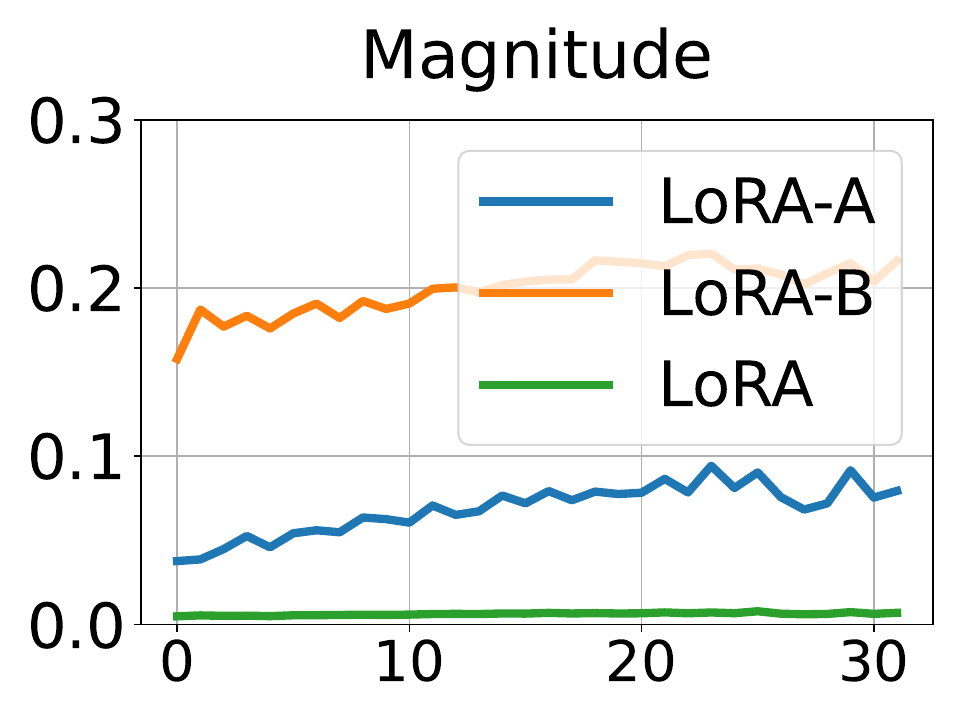}
\includegraphics[width=0.49\linewidth]{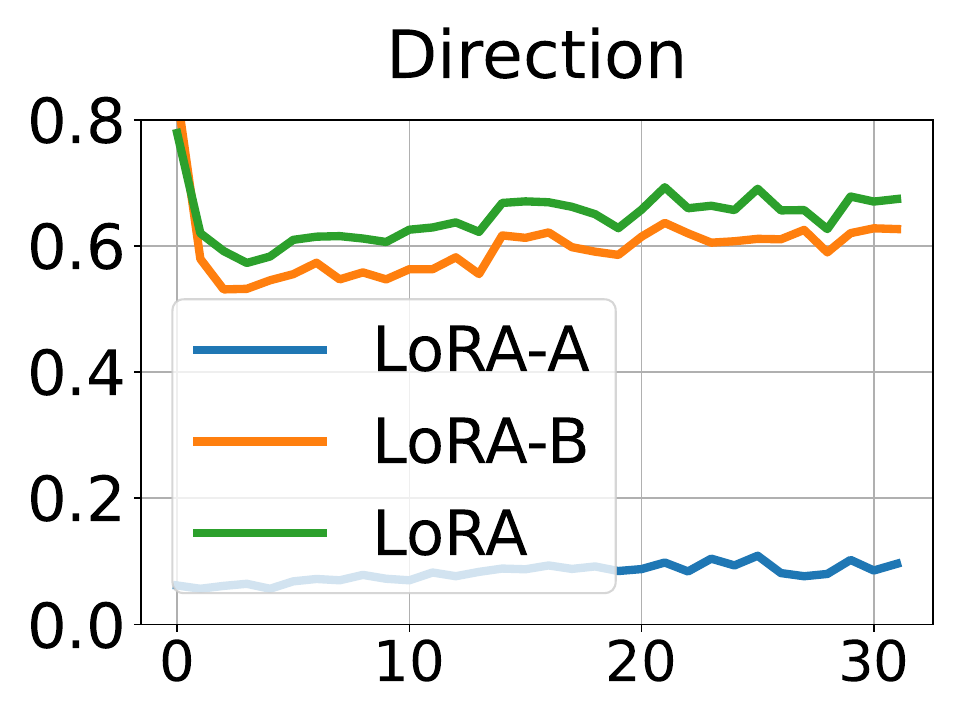}
\end{center}
\vspace{-0.5cm}
\caption{Comparison of LoRA modules variations before and after the fine-tuning. The x-axis denotes the layer indices, and the y-axis denotes the variation values. The module $B$ exhibits pronounced variation in both magnitude and direction. Overall, LoRA shows limited magnitude change, with nearly all directional change captured by $B$.}
\label{fig:fig2}
\end{figure}

\vspace{0.1cm}
\noindent{\bf Observation.} Figure~\ref{fig:fig1d} shows that $A$ remains highly similar throughout training, undergoing only minimal changes, whereas $B$ exhibits much larger differences, indicating substantial adaptation after fine-tuning. Figure~\ref{fig:fig2} further reveals that the variations in $A$ are primarily in magnitude with little directional change, while $B$ accounts for most of the direction change. We also conduct variation analysis using different models and settings and observe similar findings. The details are provided in Appendix~\ref{appx:learning_dynamics}. These results suggest that $A$ functions more as a fixed feature projector, whereas $B$ aggregates and adapts these features to encode domain knowledge. This highlights the more dominant role of $B$ over $A$ in knowledge learning, raising an intriguing question: might sharing the module $B$, rather than $A$, be more effective for parameter and knowledge sharing?

\subsection{Sharing $A$ or Sharing $B$?}\label{subsec:sharing_A_vs_B}

In this section, we address the question from Section~\ref{subsec:dynamics} by comparing the performance of sharing modules $A$ and $B$.

\begin{figure}[t]
\begin{center}
\includegraphics[width=0.49\linewidth]{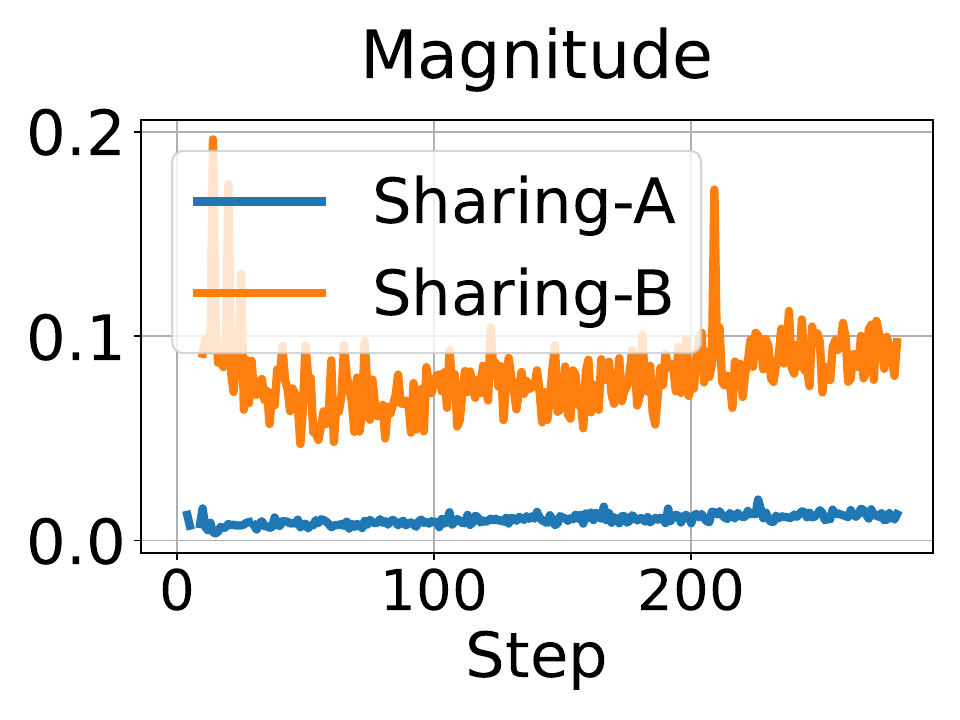}
\includegraphics[width=0.49\linewidth]{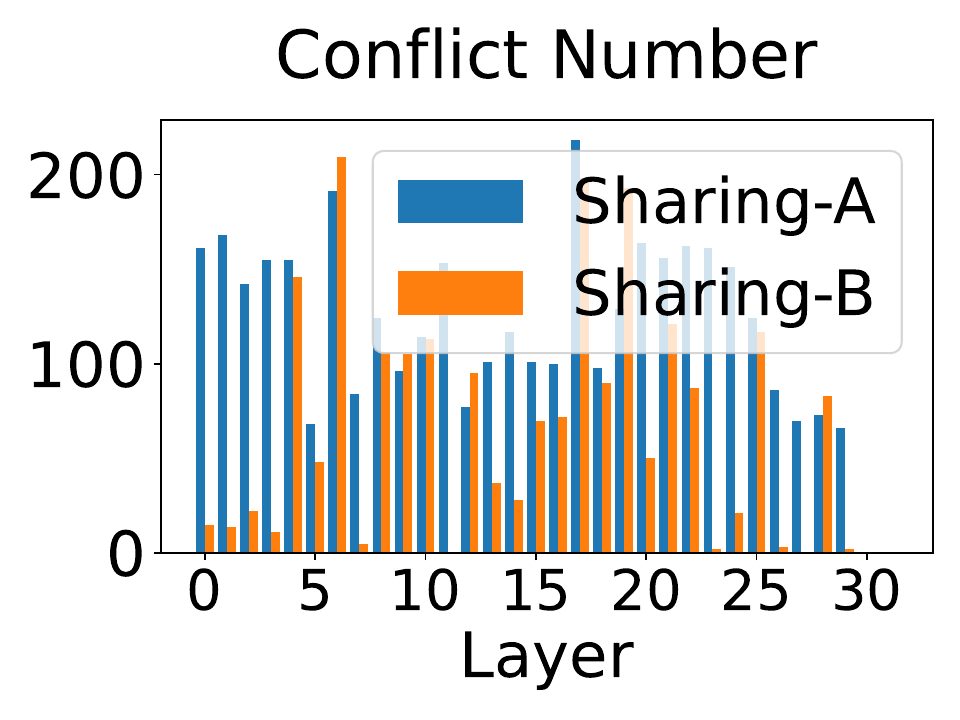}
\end{center}
\vspace{-0.5cm}
\caption{
Comparing sharing $A$ versus $B$ in multi-task fine-tuning.  
Left: gradient magnitudes of $A$ and $B$. 
Right: number of gradient conflicts per layer. 
Sharing $A$ causes smaller gradient magnitudes and more frequent conflicts than sharing $B$.}
\label{fig:fig3}
\vspace{-0.5cm}
\end{figure}

\vspace{0.1cm}
\noindent\textbf{Gradient conflicts lead to lazy learning for $A$ in multi-task fine-tuning}. Given an input $x \in \mathbb{R}^{d_{\text{in}}}$ and the output gradient $g \in \mathbb{R}^{d_{\text{out}}}$, the gradient of $A$ in sharing-$A$ structure is
$
\nabla A = \sum_{i=1}^n w_i (B_i^\top g) x^\top,
$
where each term corresponds to a $B_i$ expert. We record the magnitudes of $\nabla A$ and compute cosine similarity between gradient components, where negative similarity indicates a conflict that may hinder learning. 
We apply same procedure to the sharing-$B$ structure and compare them on commonsense reasoning \citep{hu2023llm}. We track gradient magnitudes of shared parameters and the number of conflicts during training. 

\vspace{0.1cm}
\noindent{\bf Observation.} Figure~\ref{fig:fig3}(Left) shows that the gradient magnitude of $A$ in the sharing-$A$ structure is near zero, while the gradient of $B$ in the sharing-$B$ structure is much larger. Figure~\ref{fig:fig3}(Right) shows that sharing $A$ also produces more gradient conflicts.
Thus, in the sharing-$A$ structure, $A$ learns very slowly possibly due to the more frequent conflicting updates. We refer to this phenomenon as ``lazy learning''. Previous analysis in Section~\ref{subsec:dynamics} indicates that $A$ functions as a feature projector. Hence, ``lazy learning'' may restrict the ability to explore diverse feature subspaces. 

\vspace{0.1cm}
\noindent\textbf{Knowledge transfer in federated fine-tuning}. Each client fine-tunes its own LoRA and transmits the shared parameters to the server, which aggregates and returns them (see Section~\ref{sec:fedalora}). We assess whether the shared parameters improve knowledge transfer by evaluating each client’s performance across all tasks. 
\begin{wraptable}{r}{0.18\textwidth}
\vspace{-6pt}
\centering
\small 
\resizebox{\linewidth}{!}{
\begin{tabular}{ccc}
\toprule
Setting & Sharing & Avg. \\
\midrule
\multirow{2}{*}{Homo.}
& $A$ & 44.30 \\
& $B$ & \textbf{66.32} \\
\midrule
\multirow{2}{*}{Hetero.}
& $A$ & 40.76 \\
& $B$ & \textbf{50.30} \\
\bottomrule
\end{tabular}
}
\caption{Comparing sharing $A$ and sharing~$B$ in federated fine-tuning. 
}
\label{tab:transfer}
\vspace{-6pt} 
\end{wraptable}
We compare these two structures across 8 clients, each assigned an NLP task from the FLAN dataset \citep{weifinetuned}, and use ROUGE-1 score ~\citep{lin2004rouge} to measure performance, where a value of 0 means no overlap between model prediction and ground truth, and 100 indicates perfect word-level overlap.  We report the average score across clients in Table~\ref{tab:transfer}. See Appendix~\ref{appx:extra_exps_sharing_A_B} for detailed results for each client.

\vspace{0.1cm}
\noindent{\bf Observation.}  In homogeneous setting, sharing $B$ outperforms sharing $A$ by an average of 49.71\%. In heterogeneous setting, sharing $B$ achieves an average improvement of 23.41\%. These results clearly indicate that sharing $B$ better facilitates cross-client knowledge transfer than sharing $A$.

\vspace{-0.2cm}

\section{Proposed Methods}
\vspace{-0.2cm}

Motivated by the findings in Section~\ref{sec:revisiting}, we replace $A$ with $B$ as the shared parameter and propose two simple and  effective multi-LoRA fine-tuning methods: \textbf{ALoRA} (Asymmetric LoRA) for multi-task training, and \textbf{Fed-ALoRA} for both homogeneous and heterogeneous federated settings.

\begin{figure}[t]
\begin{center}
\includegraphics[width=0.85\linewidth]{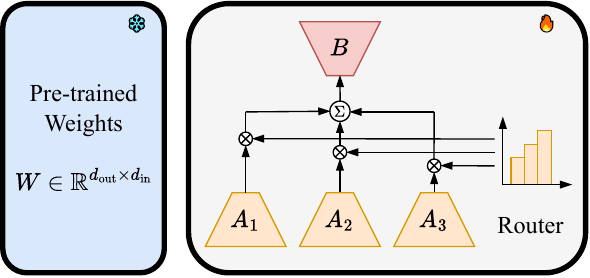}
\end{center}
\vspace{-0.3cm}
\caption{ALoRA adopts multiple $A$ and a single $B$ to explore diverse feature subspaces.}
\label{fig:fig4}
\vspace{-0.4cm}
\end{figure}

\vspace{-0.1cm}
\subsection{ALoRA for Multi-Task Fine-Tuning}
\vspace{-0.1cm}

Multi-task fine-tuning typically adapts a pretrained LLM using data from multiple tasks. The goal is to improve generalization by learning from diverse inputs. The proposed ALoRA is illustrated in Figure~\ref{fig:fig4}. Given input $x\in\mathbb{R}^{d_{\text{in}}}$, the forward pass is
\begin{align*}
    y=y_0+\Delta y=W_0x+ B\sum_{i=1}^n w_iA_ix,
    \end{align*}
where $W_0\in\mathbb{R}^{d_{\text{out}}\times d_{\text{in}}}$ is the pre-trained weight matrix, $A_i\in\mathbb{R}^{r\times d_{\text{in}}}$ are the expert matrices, $B\in\mathbb{R}^{d_{\text{out}}\times r}$ is the shared aggregator, and the rank $r\ll min(d_{\text{in}},d_{\text{out}})$. Each $A_i$ projects the input into a distinct feature subspace, and $B$ fuses the learned features to produce the output. The expert weights $w = (w_1, \ldots, w_n)$ are obtained from an input-aware router, implemented as a linear gating function with parameters $W_g \in \mathbb{R}^{n \times d_{\text{in}}}$:
   $ w=\text{softmax}(W_gx).$ During the inference, the router computes input-dependent weights, and the weighted average of the adapters is dynamically merged into the pre-trained weights.

\subsection{Fed-ALoRA for Federated Fine-Tuning}\label{sec:fedalora} 

\begin{figure*}[h]
\begin{center}
\includegraphics[width=0.85\linewidth]{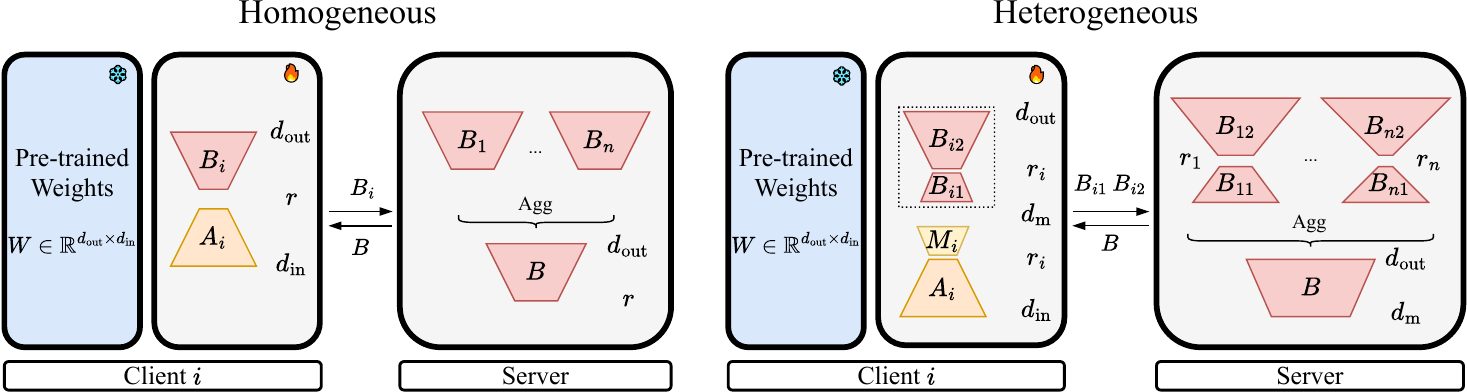}
\end{center}
\vspace{-0.2cm}
\caption{Fed-ALoRA shares only $B$ matrices for server aggregation. Left: Homogeneous setting (same rank), where the shared $B$ is directly transmitted. Right: Heterogeneous setting (different ranks), where the shared $B$ is decomposed into two matrices for heterogeneity. Compared to the standard full LoRA aggregation, the communication cost per client is reduced to $\mathcal{O}(d_\text{out}r)$ in the homogeneous setting and $\mathcal{O}(d_\text{out}r_i)$ in the heterogeneous setting if $d_m$ is chosen appropriately.}
\label{fig:fig5}
\vspace{-0.2cm}
\end{figure*}

Federated fine-tuning can be divided into two settings: (i) \emph{homogeneous}, where all clients adopt the same configuration, and (ii) \emph{heterogeneous}, where clients have varying capacities, introducing both computational and communication heterogeneity. 

\vspace{0.1cm}
\noindent\textbf{Homogeneous setting}. In this case, all $n$ clients fine-tune their LoRA modules with the same rank. Each update takes the form $\Delta W_i = B_iA_i$, where $A_i \in \mathbb{R}^{r \times d_{\text{in}}}$ and $B_i \in \mathbb{R}^{d_{\text{out}} \times r}$. Figure~\ref{fig:fig5}(Left) illustrates the procedure of Fed-ALoRA for homogeneous setting, with details for each communication round $t$ below $(t\ge1)$: 
\vspace{0.1cm}

\begin{enumerate}[leftmargin=*, label=S\arabic*:,nosep]
    \item {\em Initialization.} If $t=1$, each client initializes $A_i$ randomly and sets $B_i$ to zero. For $t>1$, $A_i$ and $B_i$  are initialized with $A_i^{t-1}$ and $B_0^{t-1}$.
    \item {\em Local training.} Each client performs LoRA fine-tuning on its local data, obtaining $(A_i^t,B_i^t)$ by optimizing $\mathcal{L}(W_0+B_iA_i)$ with respect to $(A_i,B_i)$, where $\mathcal{L}(\cdot)$ is the loss function. The client then uploads only $B_i^t$ to the server for aggregation.
    \item {\em Aggregation.} The server aggregates the uploaded matrices using the operator $\text{Agg}(\cdot)$ from \citet{mcmahan2017communication}, and obtains  $B_0^t\leftarrow \text{Agg} (B_1^t,\cdots,B_n^t).$  
    \item {\em Broadcast.} The server then sends the global matrix $B_0^t$ back to all clients.
\end{enumerate}
\vspace{0.1cm}
In full-LoRA aggregation, each client communicates $(d_{\text{in}} + d_{\text{out}})r$ parameters per round. In contrast, Fed-ALoRA requires transmitting only $B_i$, reducing communication cost to $d_{\text{out}}r$.

\vspace{0.1cm}

\noindent\textbf{Heterogeneous setting}. In this case, clients may have different capacity constraints, resulting in parameterizations with diverse ranks $r_i$, given by $\Delta W_i = B_iA_i$, where $A_i \in \mathbb{R}^{r_i \times d_\text{in}}$ and $B_i \in \mathbb{R}^{d_\text{out} \times r_i}$ for $i=1,\ldots,n$. Since the ranks $r_i$ differ across clients, direct averaging of the $B_i$ matrices is infeasible, and the aggregation strategy used in the homogeneous setting cannot be applied.   

\begin{figure}[H]
\begin{center}
\includegraphics[width=0.85\linewidth]{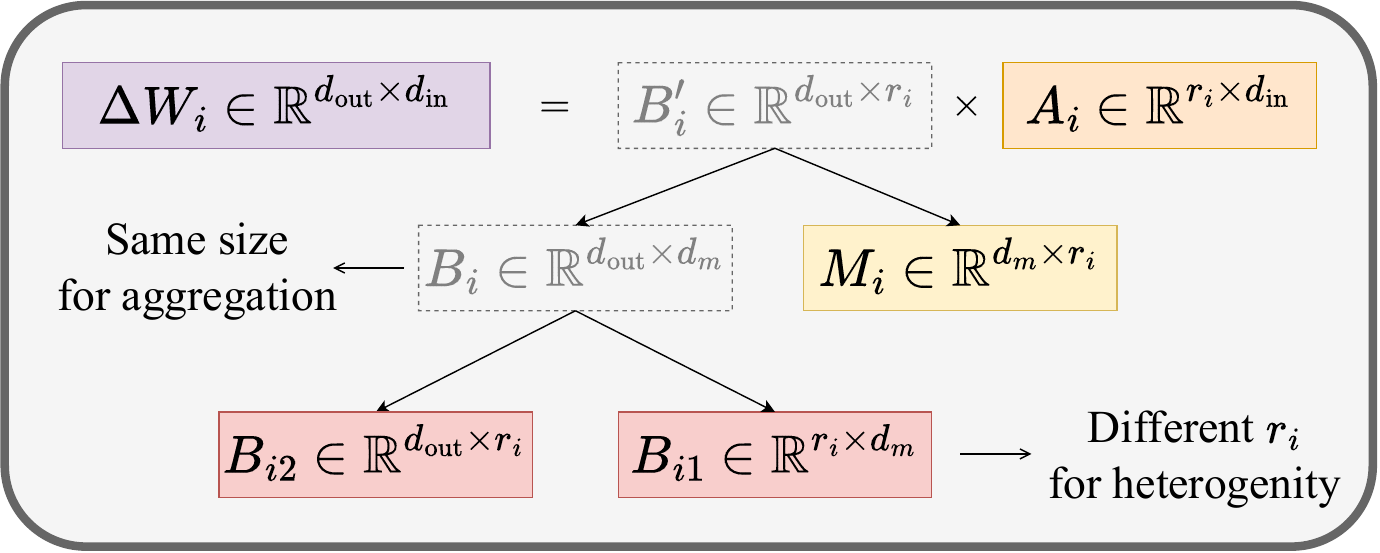}
\end{center}
\vspace{-0.2cm}
\caption{Our proposed decomposition strategy. All clients have the same $B_i$ dimensionality for aggregation, and different ranks $r_i$ to support heterogeneity. }
\label{fig:decomposition}
\end{figure}

To address this issue, we propose a novel {\em decomposition strategy} of the form:
\begin{align*}
    \Delta W_i=B_{i2}B_{i1}M_iA_i,
\end{align*}
where $A_i\in\mathbb{R}^{r_i \times d_{\text{in}}}$, $M_i\in\mathbb{R}^{d_m\times r_i}$, $B_{i1}\in\mathbb{R}^{r_i \times d_m}$ and $B_{i2}\in\mathbb{R}^{d_{\text{out}}\times r_i}$. The {\em high-level idea} is illustrated in Figure~\ref{fig:decomposition}. We first represent the low-rank updates as $B_iM_iA_i$ for $i$-th client, where $M_i$ is an intermediate matrix controlling $d_m$. This design enables us to aggregate $B_i$ across all clients. To handle the heterogeneity, we further decompose $B_i$ into two components, $(B_{i1}, B_{i2})$ with rank $r_i$.
In addition, every client maintains an accumulator $B_{i0}\in\mathbb{R}^{d_\text{out}\times d_m}$ which stores the global updates it has received so far. The full procedure of round $t$ is illustrated in Figure~\ref{fig:fig5}(Right) and detailed below:

\vspace{0.1cm}

\begin{enumerate}[leftmargin=*, label=S\arabic*:,nosep]    
    \item {\em Initialization.}  If $t=1$, each client initializes $(A_i,M_i,B_{i1})$ randomly, and sets $(B_{i0},B_{i2})$ to zero. For $t>1$, $(A_i,M_i)$ is initialized with $(A_i^{t-1},M_i^{t-1})$, $B_{i0}$ is updated by accumulation as $B_{i0}\leftarrow B_{i0}+B_0^{t-1}$, $B_{i1}$ is re-initialized randomly, and $B_{i2}$ is resets to zero.
    \item {\em Local training.} Each client performs LoRA fine-tuning on its local data, and obtains parameters $(A_i^t,M_i^t,B_{i1}^t,B_{i2}^t)$ by optimizing $\mathcal{L}(W_0+(B_{i0}+B_{i2}B_{i1})M_iA_i)$ with respect to $(A_i,M_i,B_{i1},B_{i2})$. 
    The client then uploads parameters $(B_{i1}^t,B_{i2}^t)$ to the server.
    \item {\em Aggregation.} The server reconstructs $B_i^t=B_{i2}^tB_{i1}^t$ for each client and then performs the aggregation $B_0^t\leftarrow \text{Agg}(B_1^t,\cdots,B_n^t)$.
    \item {\em Broadcast.} The server then sends the global matrix $B_0^t$ back to all clients.
\end{enumerate}

\noindent \textbf{Remark 1.} The prior parameter-sharing approach, FedSA-LoRA \citep{guoselective}, does not support the heterogeneous setting. Fed-ALoRA addresses this limitation by introducing the decomposition $(B_{i1},B_{i2})$, enabling efficient aggregation across clients with different capacities.

\noindent \textbf{Remark 2.} In full-LoRA aggregation, each client uploads $(d_\text{in}+d_\text{out})r_i$ parameters to the server. The server then extends all heterogeneous updates to the maximum rank $r_\text{max}=\max\{r_1,\cdots,r_n\}$ by padding with zeros, and broadcasts $(d_\text{in}+d_\text{out})r_\text{max}$ parameters back to every client. 
In Fed-ALoRA, if $d_m$ is chosen comparable to $r_\text{max}$ with $d_m\ll \min(d_\text{in},d_{\text{out}})$, then client $i$ maintains $(d_\text{in}+d_\text{out}+2d_m)r_i\approx (d_\text{in}+d_\text{out})r_i$ trainable parameters. The communication cost is $\mathcal{O}(d_\text{out} r_i)$ for uplink and $\mathcal{O}(d_\text{out} r_{\max})$ for downlink.

\section{Experiments}

\subsection{Multi-Task Fine-Tuning}
{\bf Models and datasets.} We fine-tune LLaMA3-8B  \citep{dubey2024llama} and Qwen2-7B \citep{yang2025qwen3} on the \emph{intra-domain} multi-task benchmark commonsense reasoning \citep{hu2023llm}, which contains 8 question answering (QA) datasets, each focusing on a different aspect of commonsense. We also fine-tune LLaMA2-7B and Qwen2-7B on the \emph{cross-domain} multi-task NLP dataset \citep{long2024dual}, which mixes 8 different tasks such as text generalization. These tasks are sampled from the FLAN dataset \citep{weifinetuned}. Additional results on the math reasoning benchmark~\citep{hu2023llm} are provided in Appendix~\ref{appx:experiments}.

\vspace{0.1cm}
\noindent{\bf Baseline methods.} For each task (or client in the federated setting), we first fine-tune LoRA on its own dataset and use the performance as the single-task  baseline, denoted as ST Baseline. We compare our ALoRA with representative methods, including vanilla LoRA \citep{hulora}, LoHa \citep{yeh2023navigating}, AdaLoRA \citep{zhangadaptive}, MoSLoRA \citep{wu2024mixture}, and HydraLoRA \citep{tian2024hydralora}. 
Although our focus is on parameter sharing in a multi-LoRA structure without relying on task specific information, such as task specific transformation matrices between the LoRA $A$ and $B$ matrices or task specific LoRA modules, we still compare our method with multi-task fine-tuning approaches that explicitly exploit such information, including LoRAMoE~\citep{dou2023loramoe}, MTL-LoRA~\citep{yang2025mtl}, and CoLA~\citep{zhou2025cola}. Our ALoRA performs better than or comparably to theses methods. Please refer to Appendix~\ref{appx:comparision_extra_mtft} for more details.

\vspace{0.1cm}
\noindent{\bf Evaluation metrics.} To evaluate performance, we use the following metrics: (1) average accuracy for commonsense reasoning and average ROUGE-1 score for multi-task NLP dataset; and (2) $\Delta m\%$ \citep{maninis2019attentive}, the average per-task performance change against the single-task baseline. $\Delta m\%=\frac{1}{K}\sum_{k=1}^K (-1)^{\delta_k} (M_k-M_0)/M_0\times 100$, where $M_k$ is the performance of $k$-th task under the compared method, $M_0$ is the baseline performance. $\delta_k=1$ if higher values indicate better performance, otherwise $\delta_k=0$. This metric evaluates how well performance is balanced across multiple tasks.

\begin{table}[t]
\centering
\begin{adjustbox}{width=0.95\linewidth}
\begin{tabular}{c|cc|cc}
\toprule\toprule
\multirow{2}{*}{Method}
& \multicolumn{2}{c|}{LLaMA3-8B}
& \multicolumn{2}{c}{Qwen2-7B} \\
\cmidrule(lr){2-3}\cmidrule(lr){4-5}
& Avg. & $\Delta m\%(\downarrow)$
& Avg. & $\Delta m\%(\downarrow)$ \\
\midrule\midrule

ST Baseline
& 84.90 & --   
& 87.27 & -- \\

LoRA
& 83.64 & 1.48
& 85.96 & 1.43 \\
LoHa
& 83.73 & 1.36
& 86.34 & 1.06 \\
AdaLoRA
& 83.88 & 1.17
& 86.14 & 1.30 \\
MoSLoRA
& 84.23 & 0.76
& 86.30 & 1.09 \\
HydraLoRA
& 84.57 & 0.32
& 86.09 & 1.32 \\
\rowcolor{oursblue}
ALoRA (Ours)
& \textbf{84.81} & \textbf{0.04}
& \textbf{86.47} & \textbf{0.91} \\
\bottomrule\bottomrule
\end{tabular}
\end{adjustbox}
\vspace{-0.2cm}
\caption{Results on intra-domain multi-task commonsense reasoning benchmark. $\Delta m\%$ measures performance balance across tasks. $\downarrow$ denotes that lower values are better.  All methods use the same number of adapter parameters. We run each experiment 3 times and report the average. Full details, including per-task performance and standard deviations, are provided in Appendix~\ref{appx:extra_exps_commonsense}.}
\label{tab:tab2}
\end{table}

\noindent{\bf Results.} The results are presented in Tables~\ref{tab:tab2} and~\ref{tab:tab3}. ALoRA achieves better average scores than existing LoRA variants with the most balanced results in both benchmarks. 
These results suggest that ALoRA encourages knowledge transfer.

\begin{table}[t]
\centering
\begin{adjustbox}{width=0.95\linewidth}
\begin{tabular}{c|cc|cc}
\toprule\toprule
\multirow{2}{*}{Method}
& \multicolumn{2}{c|}{LLaMA2-7B}
& \multicolumn{2}{c}{Qwen2-7B} \\
\cmidrule(lr){2-3}\cmidrule(lr){4-5}
& Avg. & $\Delta m\%(\downarrow)$
& Avg. & $\Delta m\%(\downarrow)$ \\
\midrule\midrule

ST Baseline

& 63.36 & --    
& 76.16 & -- \\
LoRA
& 61.67 & 0.31
& 79.60 & -6.78 \\
LoHa
& 65.70 & -5.76
& 78.35 & -5.27 \\
AdaLoRA
& 61.96 & -0.41
& 77.61 & -4.33 \\
MoSLoRA
& 66.18 & -6.58
& 79.34 & -6.59 \\
HydraLoRA
& 66.45 & -6.39
& 80.03 & -7.14 \\
\rowcolor{oursblue}
ALoRA (Ours)
& \textbf{67.13} & \textbf{-8.33}
& \textbf{80.46} & \textbf{-7.98} \\
\bottomrule\bottomrule
\end{tabular}
\end{adjustbox}
\vspace{-0.2cm}
\caption{Results on cross-domain multi-task NLP datasets. $\Delta m\%$ measures performance. Full details on per-task performance are provided in Appendix~\ref{appx:extra_exps_mtl_nlp}.}
\label{tab:tab3}
\vspace{-0.2cm}
\end{table}

\subsection{Federated Fine-Tuning}

\noindent{\bf Models, datasets and metrics.} We fine-tune models LLaMA2-7B and Qwen2-7B, and evaluate on the federated datasets constructed by \citet{long2024dual}, which includes 8 NLP tasks from FLAN dataset \citep{weifinetuned}, with each client assigned to one task. We use ROUGE-1 score~\citep{lin2004rouge} as the evaluation metric. We also report the average number of parameters (in millions) communicated per client in each round, including both uploads to the server and downloads from the server.

\vspace{0.1cm}
\noindent{\bf Baseline methods.}
We compare our Fed-ALoRA with  state-of-the-art federated fine-tuning approaches including FedIT \citep{zhang2024towards}, FedDPA \citep{long2024dual}, FedSA-LoRA \citep{guoselective}, ZeroPadding~\citep{wang2024flora}, and FLoRA \citep{wang2024flora}. 
 In the homogeneous setting, all clients use rank 8.  In the heterogeneous setting, the ranks are $\{64,64,32,32,16,16,8,8\}$, with $d_m$ set to 16. Full details are provided in the Appendix~\ref{appx:experiments}.

\vspace{0.1cm}
\noindent{\bf Results.} The results are presented in Tables~\ref{tab:tab4} and \ref{tab:tab5}. Fed-ALoRA achieves the highest average score and most balanced performance while reducing communication cost by 50\% compared to full-LoRA aggregation in homogeneous setting, and reducing by 75\% in heterogeneous setting. 
These results show that Fed-ALoRA effectively promotes knowledge sharing across clients.

\begin{table}[t]
\centering
\begin{adjustbox}{width=1\linewidth}
\begin{tabular}{c|ccc|ccc}
\toprule\toprule
\multirow{2}{*}{Method}
& \multicolumn{3}{c|}{LLaMA2-7B}
& \multicolumn{3}{c}{Qwen2-7B} \\
\cmidrule(lr){2-4}\cmidrule(lr){5-7}
& Avg. & $\Delta m\%(\downarrow)$ & Params.
& Avg. & $\Delta m\%(\downarrow)$ & Params. \\
\midrule\midrule

ST Baseline

& 79.67 & --    & --
& 82.75 & --    & -- \\
FedIT
& 82.47 & -3.92 & 8.39
& 82.80 & -0.05 & 6.42 \\
FedDPA
& 81.96 & -3.30 & 16.78
& 83.17 & -0.60 & 12.85 \\
FedSA-LoRA
& 81.15 & -2.21 & \textbf{4.19}
& 83.69 & -1.15 & \textbf{3.21} \\
\rowcolor{oursblue}
Fed-ALoRA (Ours)
& \textbf{82.51} & \textbf{-4.29} & \textbf{4.19}
& \textbf{84.30} & \textbf{-2.05} & \textbf{3.21} \\
\bottomrule\bottomrule
\end{tabular}
\end{adjustbox}
\vspace{-0.2cm}
\caption{Results for the {\bf homogeneou}s federated setting. Params. denotes the average number of parameters (in millions) transmitted per client in each round. ALoRA achieves the most balanced performance while reducing communication cost by 50\% compared to full LoRA aggregation FedIT. Full details on per-client performance are provided in Appendix~\ref{appx:extra_exps_homo}.}
\label{tab:tab4}
\vspace{-0.1cm}
\end{table}

\begin{table}[t]
\centering
\begin{adjustbox}{width=\linewidth}
\begin{tabular}{c|ccc|ccc}
\toprule\toprule
\multirow{2}{*}{Method}
& \multicolumn{3}{c|}{LLaMA2-7B}
& \multicolumn{3}{c}{Qwen2-7B} \\
\cmidrule(lr){2-4}\cmidrule(lr){5-7}
& Avg. & $\Delta m\%(\downarrow)$ & Params.
& Avg. & $\Delta m\%(\downarrow)$ & Params. \\
\midrule\midrule
ST Baseline

& 81.73 & --    & --    
& 83.43 & --    & -- \\
ZeroPadding
& 82.29 & -0.91 & 49.28
& 83.32 & 0.06  & 37.73 \\
FLoRA
& 80.45 & 1.54  & 141.56
& 81.65 & 2.13  & 108.38 \\
\rowcolor{oursblue}
Fed-ALoRA (Ours) & \textbf{82.50} & \textbf{-1.07} & \textbf{12.12} & \textbf{84.13} & \textbf{-1.02} & \textbf{9.23} \\
\bottomrule\bottomrule
\end{tabular}
\end{adjustbox}
\vspace{-0.2cm}
\caption{Results for the {\bf heterogeneous} setting. ALoRA achieves the most balanced performance while reducing communication cost by 75\% compared to full LoRA aggregation ZeroPadding. FedSA-LoRA is not included here because it does not support heterogeneity. Full details on per-client performance are provided in Appendix~\ref{appx:extra_exps_hetero}.}
\label{tab:tab5}
\vspace{-0.4cm}
\end{table}

\subsection{In-Depth Analysis}\label{exp:depth_analysis}

\begin{figure}[t]
\begin{center}
\includegraphics[width=0.495\linewidth]{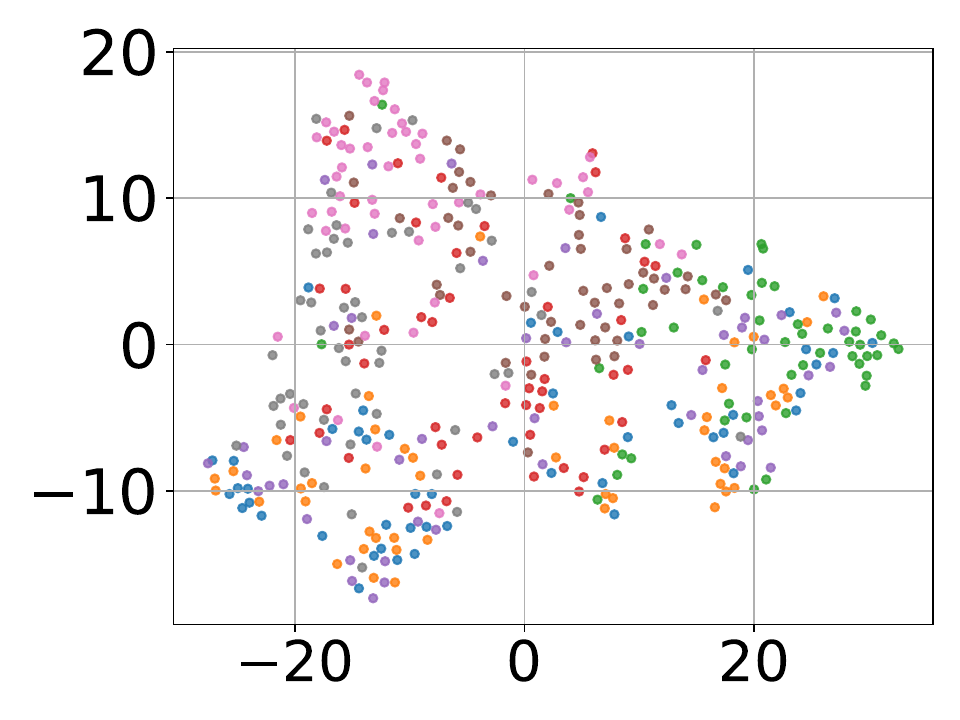}
\includegraphics[width=0.49\linewidth]{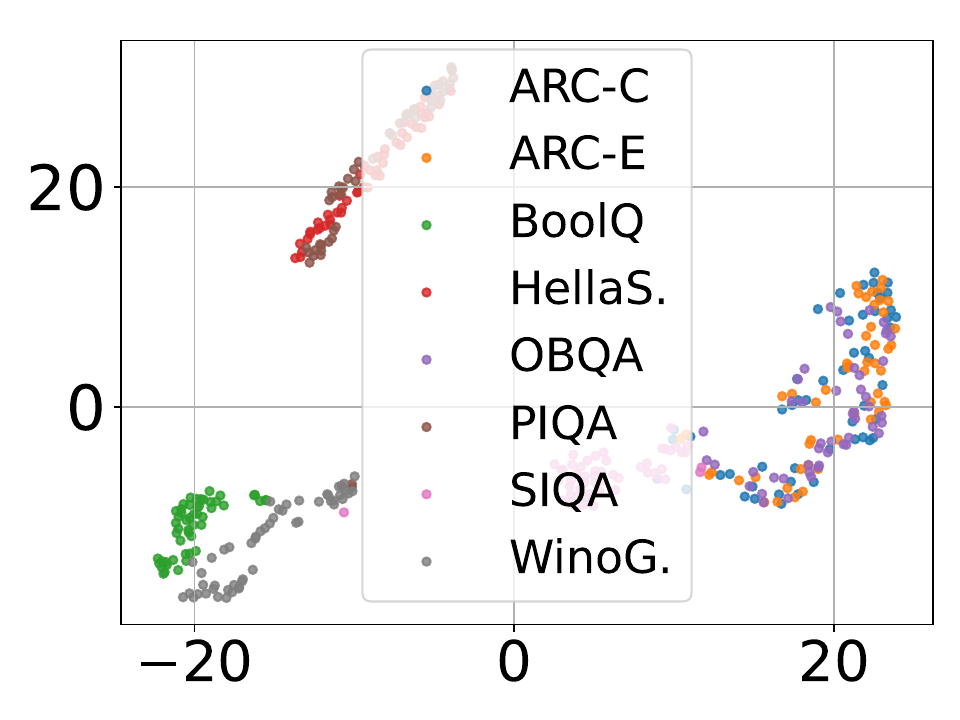}
\end{center}
\vspace{-0.3cm}
\caption{Gate activations of HydraLoRA (Left) and ALoRA (Right). ALoRA yields more distinct clusters.}
\label{fig:fig6}
\vspace{-0.3cm}
\end{figure}

\textbf{Multi-task fine-tuning}. We analyze gate activations of HydraLoRA and ALoRA during inference on the commonsense reasoning using LLaMA3-8B. Figure~\ref{fig:fig6} presents the t-SNE visualization in the last layer. ALoRA forms clearer clusters than HydraLoRA. Some tasks share similar gate activations, suggesting that they prefer same $A$ experts. This indicates that ALoRA captures diverse feature subspaces more effectively. In contrast, HydraLoRA relies on a single $A$ matrix, which limits diversity and leads to more scattered activations.

We also compare HydraLoRA and ALoRA on commonsense reasoning tasks using Qwen2.5-14B \citep{yang2025qwen3} to further validate our method. ALoRA achieves an average accuracy of 89.59, while HydraLoRA achieves 89.50. Full details are provided in Appendix~\ref{appx:comparison_large_scale}. These results show that sharing $B$ consistently outperforms sharing $A$ as the model size increases.

\vspace{0.1cm}

\noindent\textbf{Federated fine-tuning}. We study the effect of different choices of the intermediate rank $d_m$ in the heterogeneous setting using LLaMA2-7B model.
\begin{wraptable}{r}{0.25\textwidth}
\vspace{-6pt}
\small
\centering
\resizebox{\linewidth}{!}{
\begin{tabular}{ccc}
\toprule
    Fed-ALoRA & Avg. & $\Delta m\%(\downarrow)$\\
    \midrule
    $d_m=8$ & 81.92 & -0.41 \\
    $d_m=16$ & \textbf{82.50} & \textbf{-1.07} \\
    $d_m=32$ & 82.25 & -0.80 \\
    $d_m=64$ & 82.37 & -1.01 \\
    \bottomrule
\end{tabular}
}
\caption{Results of different intermediate ranks in the heterogeneous setting.}
\label{tab:immediate_rank}
\vspace{-0.2cm}
\end{wraptable}
The results in Table~\ref{tab:immediate_rank} show that with a proper choice of the rank $d_m$, we can reduce the communication cost while maintaining the performance balance.

\section{Related Work}

\textbf{Low-rank adaptation}. Vanilla LoRA \citep{hulora} reparameterizes weight updates using low-rank matrices, enabling efficient fine-tuning without extra inference latency. Extensions develop along three directions. For rank allocation, AdaLoRA \citep{zhangadaptive} prunes less important singular values, and DyLoRA \citep{valipour2023dylora} trains LoRA blocks with different ranks for flexible inference. For memory efficiency, QLoRA \citep{dettmers2023qlora} applies 4-bit quantization, and  SparseLoRA \citep{khakisparselora} updates only a sparse subset of parameters using SVD. For structural variation, LoHa and LoKr \citep{yeh2023navigating} adopt Hadamard and Kronecker decompositions, and DoRA \citep{liu2024dora} separates magnitude and direction. This paper provides a deep investigation into the training dynamics of modules $A$ and $B$, demonstrating the more dominant role of $B$ in knowledge learning and transfer. 

\vspace{0.1cm}
\noindent \textbf{Multi-task fine-tuning}. Fine-tuning on multiple tasks improves generalization and transfer. A popular idea is to integrate LoRA with MoE, where experts specialize in different tasks. Among them, LoRAMoE \citep{dou2024loramoe}, MoELoRA \citep{luo2024moelora} and MoRE \citep{zhang2025more} align experts with task information to balance performance. SMoRA~\citep{zhao2025each} treats each rank as an expert, and ThanoRA~\citep{liang2025thanora} builds task-aware LoRA modules. DynMoLE \citep{li2025dynmole} uses entropy-based routing, and HydraLoRA \citep{tian2024hydralora} improves parameter efficiency by sharing $A$ matrices. In contrast to HydraLoRA, our proposed \textbf{ALoRA} shares matrices $B$, which promotes diverse feature projections and facilitates more effective knowledge transfer.

\vspace{0.1cm}
\noindent\textbf{Federated fine-tuning}. 
Models are adapted across clients while preserving data privacy. Existing methods fall into homogeneous and heterogeneous settings. In homogeneous case, FedIT \citep{zhang2024towards} aggregates full LoRA parameters, while FedSA-LoRA \citep{guoselective} reduces communication by sharing only $A$ matrices. In heterogeneous case, HetLoRA \citep{cho2024heterogeneous} supports varying ranks via self-pruning with sparse aggregation, Ravan \citep{raje2025ravan} introduces multi-head LoRA updates, and FedALT \citep{bian2025fedalt} employs MoE-based adapters; FLoRA \citep{wang2024flora} provides a unified stacking framework. 


As we were preparing the final draft of this paper, we became aware of a concurrent work, MASA~\citep{dong2025masa}, which was posted on arXiv around the same time. MASA also explores a structure with multiple $A$ matrices and a single $B$ matrix. However, there are several key differences between MASA and our method. 
(i) \emph{Motivation}. MASA is motivated by the observation that using a single $A$ matrix can create an information bottleneck, whereas our work studies which parameters should be shared to facilitate effective knowledge transfer. 
(ii) \emph{Architecture design}. MASA shares multiple $A$ matrices across adjacent layers, while we adopt a simpler design in which each layer has its own adapted modules. 
(iii) \emph{Application}. While both methods address multi task fine tuning, our approach is further extended to federated fine tuning with substantial modifications and significantly reduced communication.


\vspace{-0.1cm}

\section{Conclusion}


Our study shows that the similarity of LoRAs' $A$ matrices arises mainly from initialization rather than shared knowledge, with $B$ serving as the key component for knowledge transfer. Building on this insight, we propose ALoRA and Fed-ALoRA, which share $B$ for multi-task and federated fine-tuning. Experiments across diverse benchmarks demonstrate that these methods achieve more balanced performance while maintaining or improving accuracy over existing multi-LoRA approaches. 
Future work will further examine the distinct learning dynamics of $A$ and $B$ and develop new fine-tuning strategies inspired by these insights.

\section*{Limitations}


In this paper, we primarily focus on multi-task and federated fine-tuning settings. Our ALoRA framework adopts a simple mixture-of-experts (MoE) structure; while effective, this design choice is not exhaustive, and other more advanced and complex MoE architectures or parameter-routing mechanisms may further improve performance or efficiency. Moreover, although our experiments cover representative language modeling tasks, some important scenarios remain unexplored, such as visual instruction tuning and other multimodal or domain-specific adaptation settings.

In addition, the relationship between adaptation parameters and knowledge sharing across tasks or clients is not yet fully understood. In particular, it remains unclear whether there exists an optimal sparse or structured parameter-sharing pattern that balances specialization and generalization in low-rank adaptation. A deeper theoretical and empirical understanding of these mechanisms would be crucial for fully characterizing the behavior and limitations of low-rank adaptation methods. We leave these directions for future work.

\section*{Ethical Statement}
All our experiments use publicly available datasets and open-source pretrained models and are intended only for research purposes. The main risks stem from dataset biases and limitations of pretrained models in real-world applications. We do not advocate deployment without human oversight.


\bibliography{custom}

\appendix

\section{Additional Related Work}\label{appx:related_work}

\textbf{Parameter-efficient fine-tuning}. PEFT adapts large language models (LLM) by training only a small subset of parameters, significantly reducing the computation cost while maintaining the performance compared to full fine-tuning. Existing PEFT methods are typically categorized into the following groups. (1) Adapter-based methods, which insert small trainable modules either serially or in parallel with Transformer blocks while keeping the pre-trained model frozen \citep{houlsby2019parameter,hetowards,lei2023conditional,pfeiffer2021adapterfusion}. (2) Soft prompt-based methods, such as prefix tuning \citep{li2021prefix,zhang2023towards} and prompt tuning \citep{lester2021power,wang2023aprompt}, which prepend learnable embeddings to the input without modifying model weights. (3) Low rank adaptation (LoRA) methods \citep{hulora,dettmers2023qlora,liu2024dora}, which approximately reparameterize weight updates using low-rank matrices. These updates can be merged into frozen weights, thereby introducing no additional inference latency.

\noindent\textbf{Multi-task learning}. The primary goal of MTL is to promote positive knowledge transfer while reducing negative interference across tasks \citep{zhang2021survey,lin2023libmtl,chen2025gradient}. Existing methods are commonly divided into three categories. (1) Soft-parameter sharing, which encourages task parameters to be similar but not identical, allowing flexible knowledge transfer while maintaining distinctions. \citet{ruder2019latent} propose learning a latent meta-architecture for multiple tasks, where task layers are regularized by their distance to the latent model. (2) Modularity-based methods, which employ task-specific modules to mitigate negative transfer. For example, \citet{lin2025mtmamba++} integrate task-specific modules to avoid gradient conflicts. (3) Task-balancing optimization methods adjust task weights or gradients using loss or gradient information. These methods include task loss weighting \citep{kendall2018multi,lin2022reasonable,xiao2025ldc} and gradient weighting approaches, such as MGDA \citep{desideri2012multiple,sener2018multi} and its variants \citep{liu2021conflict,xiao2023direction}, which apply multi-objective optimization to explore Pareto-optimal solutions for balancing gradient conflicts. Other methods examine MTL from the perspective of label noise \citep{herobust}, fairness \citep{navon2022multi, ban2024fair} or analyze the sharpness of the loss landscape \citep{phan2025beyond, ban2025samo}.

\noindent\textbf{Federated learning}. FL allows multiple clients to train collaboratively in a decentralized setting while preserving data privacy by keeping raw data local. In each communication round, clients compute local updates on their private data and send only these updates (e.g., weight deltas or gradients) to a server, which aggregates them into a new global model. Existing methods are typically grouped into three categories. (1) Aggregation methods, which improve how client updates are combined, such as robust aggregation or matched averaging \citep{morafah2023flis,huang2023achieving,liu2025cpfedavg}. (2) Communication-efficient methods, which reduce the cost of transmitting updates through techniques like sparsification, or partial updates \citep{jia2025comprehensive,li2025bgefl}. (3) Personalization methods, which adapt the global model to each client’s unique data distribution, using approaches such as model splitting, or client-specific parameters \citep{diaoheterofl,deng2024fedasa,wang2025federated}.

\section{Additional Details for Section~\ref{sec:revisiting}}
\label{appx:multilora}

\subsection{Similarity Metric in LoRA Modules}\label{appx:similarity}
LoRA represents the weight updates as $\Delta W=BA$ with $A\in\mathbb{R}^{r\times d_\text{in}}$, $B\in\mathbb{R}^{d_\text{out}\times r}$. For any invertible $R\in\mathbb{R}^{r\times r}$, we have
\begin{align*}
    \Delta W=BA=(BR)(R^{-1}A).
\end{align*}
This shows that $A$ and $B$ are not individually unique. They can be arbitrarily rotated within the rank-$r$ subspace without changing $\Delta W$. As a result, directly computing the cosine similarity between $A$ or $B$ matrices can give misleading results. 

Different seeds or initializations may lead to very different $A$ and $B$, but the subspaces they span are rotation-invariant. If the subspaces align, the modules are functionally aligned. Therefore, we use the subspace similarity proposed by \citet{zhu2013angles}. Specifically, given two matrices $M_1,M_2\in\mathbb{R}^{d\times r}$, we compute SVD of each and obtain the orthonormal bases of their column spaces, $U_1,U_2\in\mathbb{R}^{r\times r}$. The similarity is then defined as
\begin{align*}
    \text{Sim}(M_1,M_2)=\frac{1}{r}\|U_1^\top U_2\|^2_F\in[0,1],
\end{align*}
where a higher value indicates a stronger alignment.

\subsection{Analysis in Dynamics of LoRA Modules}\label{appx:learning_dynamics}
\begin{figure}[h]
\vspace{-0.2cm}
\begin{center}
\includegraphics[width=0.48\linewidth]{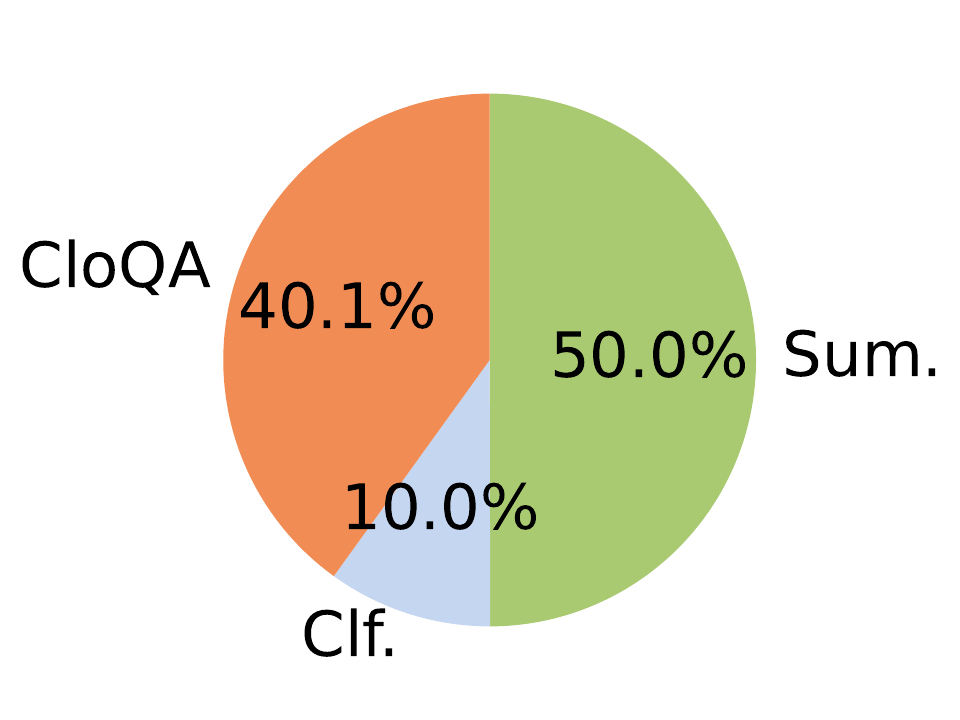}
\hspace{-1mm}
\includegraphics[width=0.48\linewidth]{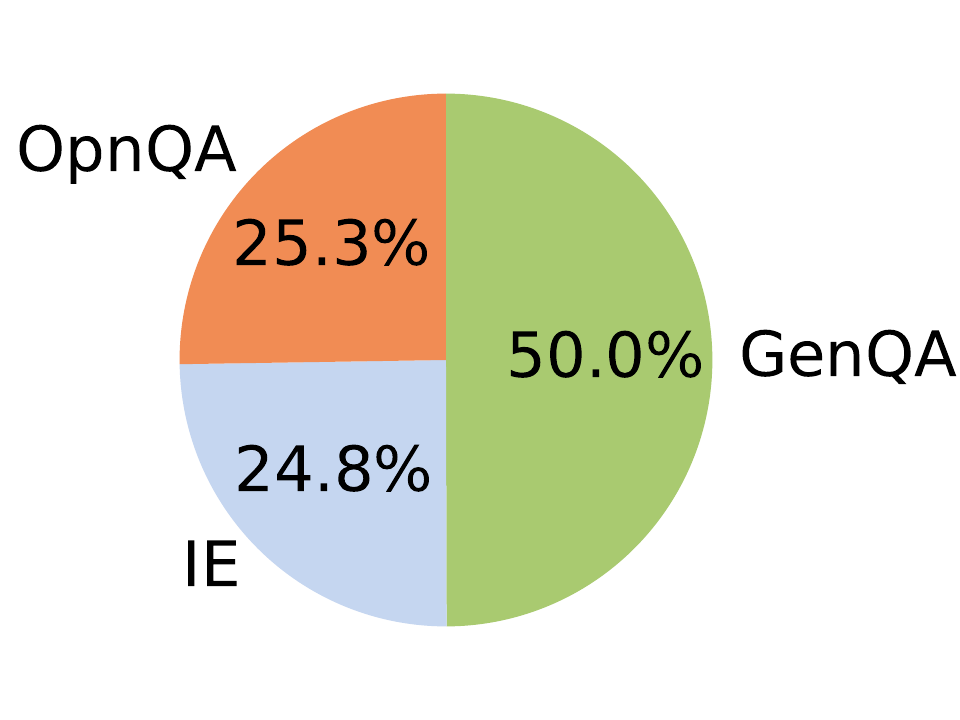}
\end{center}
\vspace{-4mm}
\caption{Data distribution of clients. Left: Client 1 contains Closed QA, Summarization, and Classification tasks. Right: Client 2 contains Open QA, General QA, and Information Extraction tasks.}
\label{fig:client_data_dist}
\vspace{-0.2cm}
\end{figure}

Section~\ref{subsec:similarity}-\ref{subsec:dynamics} analyze the learning behavior of $A$ and $B$ matrices during fine-tuning on different tasks based on LLaMA2-7B. We also perform the same analysis in the federated fine-tuning with two clients. Following \citet{zhang2024towards}, we randomly sample data from the Dolly-15K dataset \citep{DatabricksBlog2023DollyV2} and split them into two clients, each containing 1493 instruction data samples from different tasks. The data distribution is shown in Figure~\ref{fig:client_data_dist}.

\begin{figure}[t]
\centering

\begin{subfigure}[t]{0.49\linewidth}
  \centering
  \includegraphics[width=\linewidth]{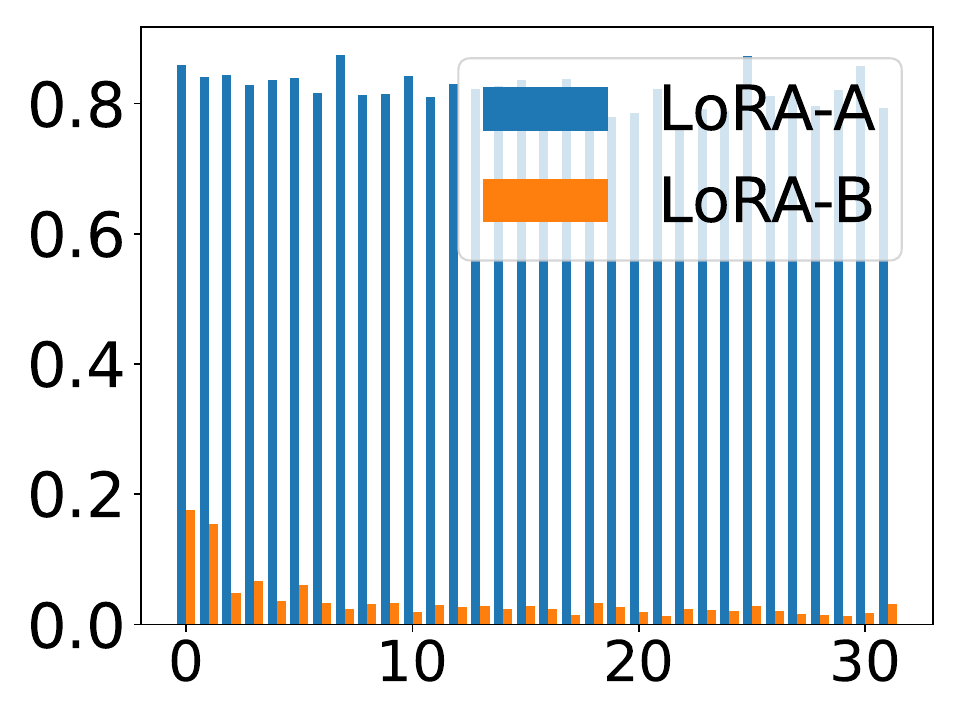}
  \caption{Different clients, same init.}
  \label{fig:fig8a}
\end{subfigure}\hfill
\begin{subfigure}[t]{0.49\linewidth}
  \centering
  \includegraphics[width=\linewidth]{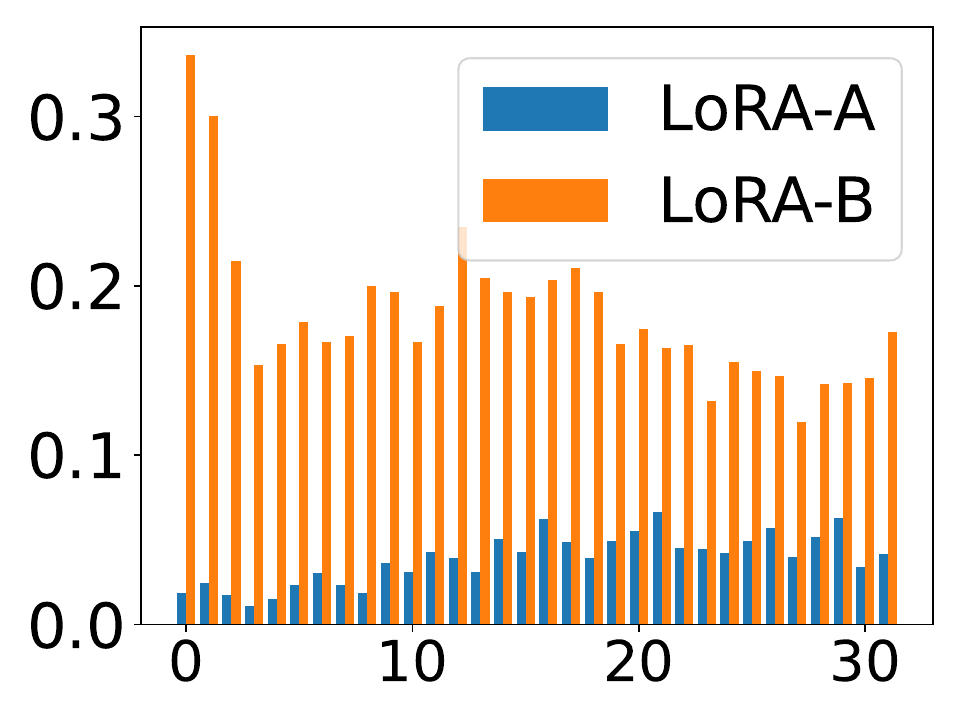}
  \caption{Same client, different init.}
  \label{fig:fig8b}
\end{subfigure}
\begin{subfigure}[t]{0.49\linewidth}
  \centering
  \includegraphics[width=\linewidth]{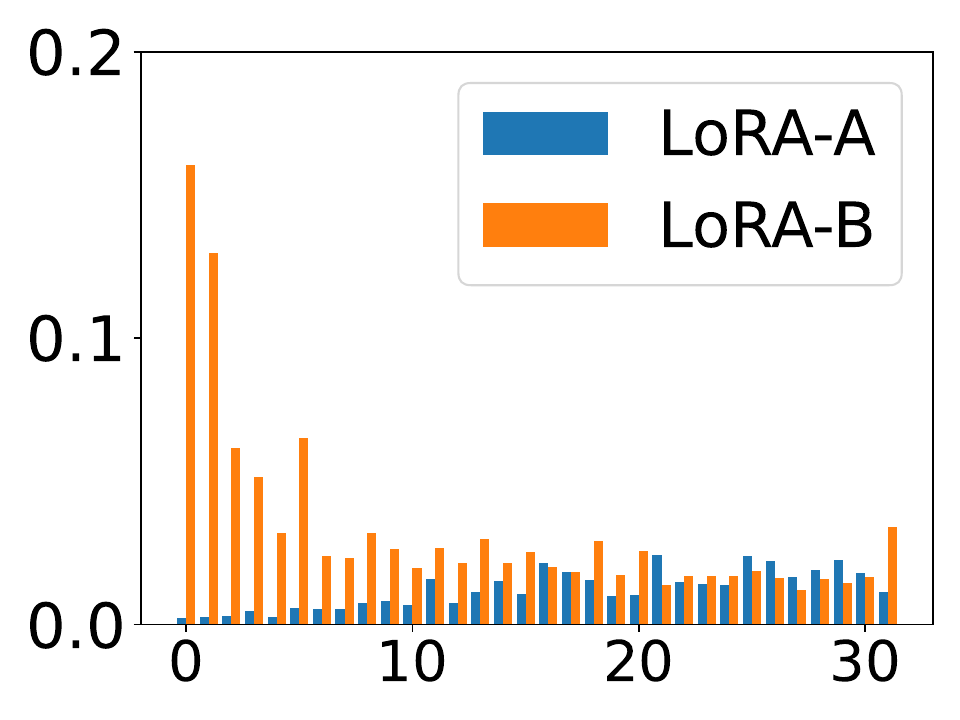}
  \caption{Different clients and init.}
  \label{fig:fig8c}
\end{subfigure}\hfill
\begin{subfigure}[t]{0.49\linewidth}
  \centering
  \includegraphics[width=\linewidth]{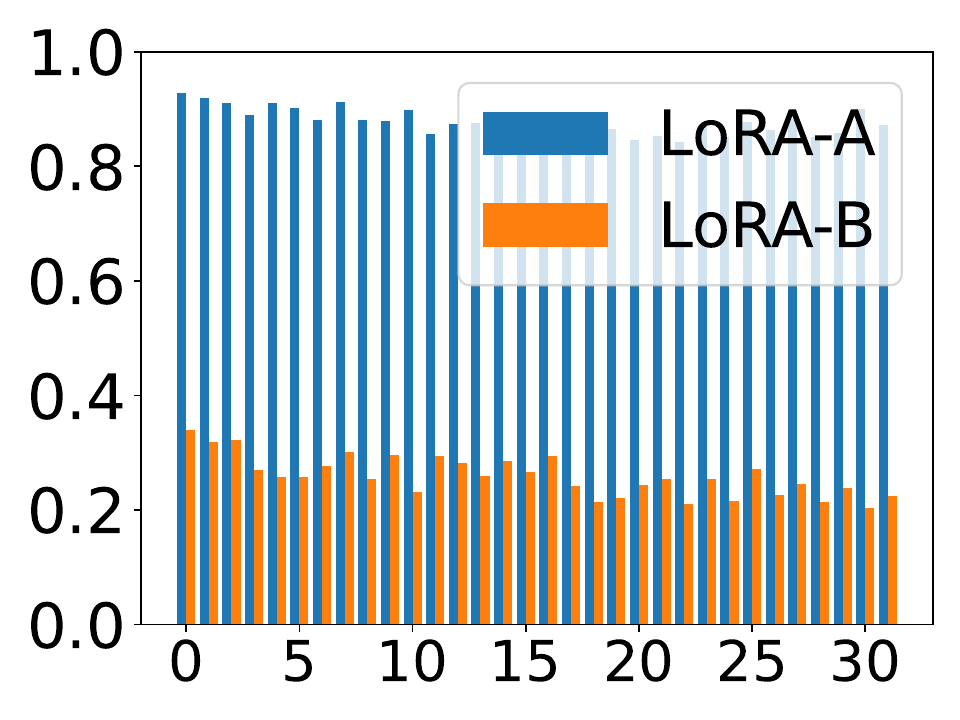}
  \caption{Before and after FT.}
  \label{fig:fig8d}
\end{subfigure}
\vspace{-0.1cm}
\caption{Layer-wise similarity analysis. The x-axis denotes the layer indices, and the y-axis denotes the similarity scores. Subfigures (a)-(c) compare different LoRA modules under clients and initialization: (a) two different clients with the same initialization; (b) the same client with different initializations; and(c) two different clients with different initializations. Subfigure (d) compares the same LoRA module before and after fine-tuning. $A$ matrices are similar only under the same initialization, whereas $B$ exhibits relatively stable similarity across different tasks and seeds. In addition, $A$ remains largely unchanged from initialization.}
\label{fig:fig8}
\end{figure}

We fine-tune the LLaMA2-7B model on the two clients and analyze how the similarity of LoRA modules is affected by random seeds for initialization. Using the similarity metric described in Appendix~\ref{appx:similarity}, we compute the layer-wise similarity of LoRA modules across the two clients. The results, shown in Figure~\ref{fig:fig8a}-\ref{fig:fig8c}, indicate that, contrary to the assumption in FedSA-LoRA \citep{guoselective}, the similarity of $A$ matrices across clients comes mainly from identical initialization rather than shared knowledge.

\begin{figure}[h]
\begin{center}
\includegraphics[width=0.49\linewidth]{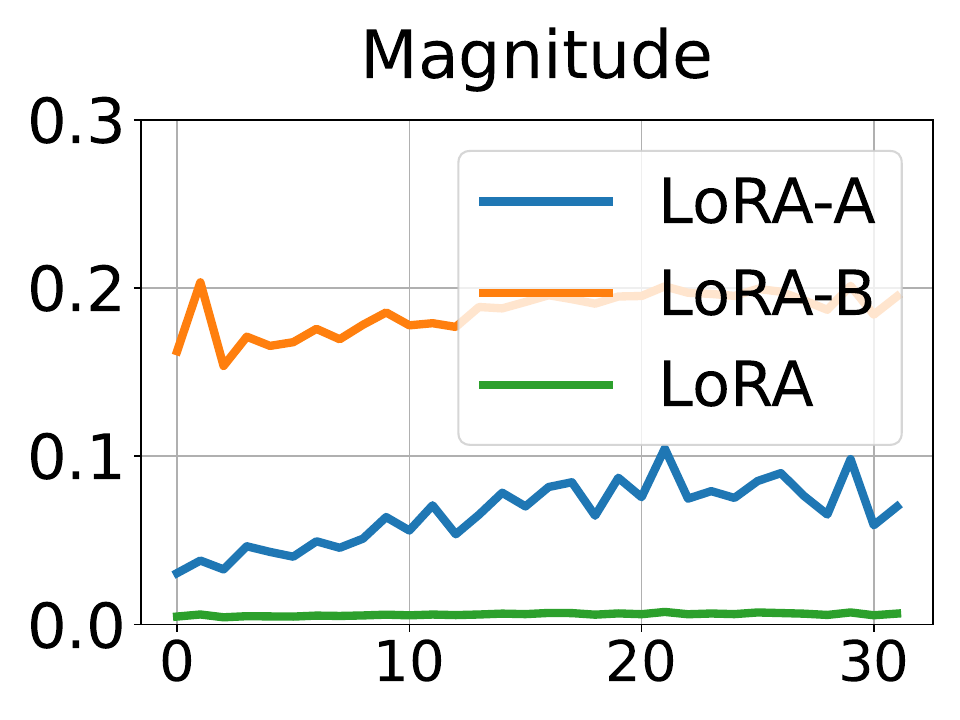}
\hspace{-1mm}
\includegraphics[width=0.49\linewidth]{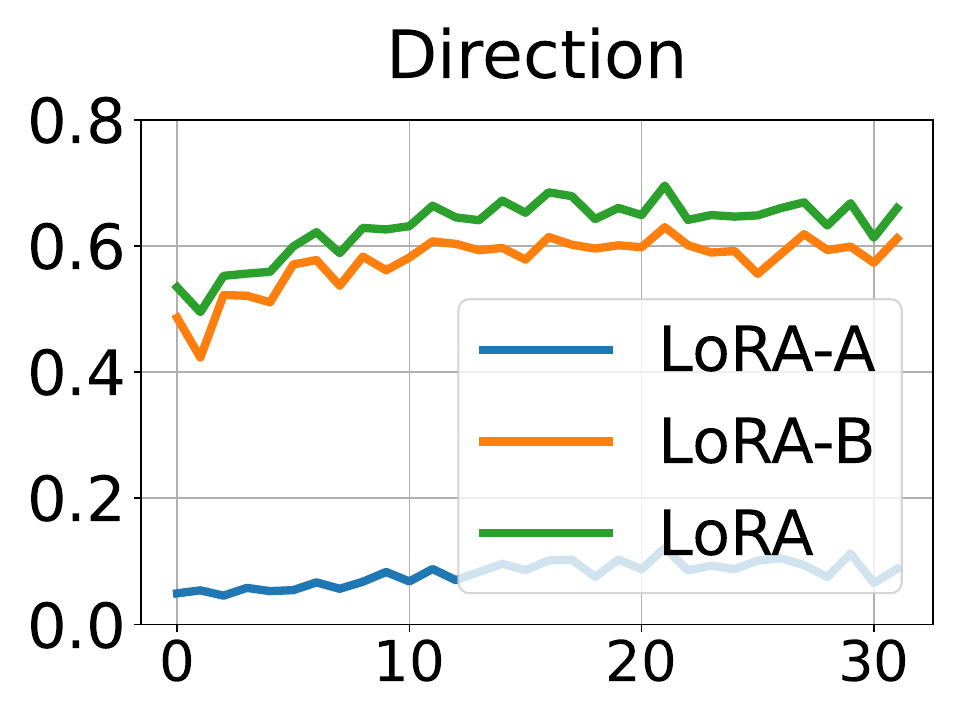}
\end{center}
\vspace{-4mm}
\caption{Comparison of LoRA modules on each client before and after federated fine-tuning. The x-axis denotes the layer indices, and the y-axis denotes the variation values. Left: magnitude change; Right: direction change. LoRA shows a limited magnitude change, with nearly all directional change captured by $B$.}
\label{fig:fig9}
\vspace{-0.4cm}
\end{figure}

Furthermore, we analyze the learning dynamics of $A$ and $B$ in the federated fine-tuning setting. On client 2, we compute the similarity of $A$ before and after fine-tuning, and do the same for $B$. We also calculate the magnitude and direction variation for this client. The results, shown in Figure~\ref{fig:fig8d}-\ref{fig:fig9}, are consistent with our earlier fine-tuning experiments on different tasks. They confirm that $B$ plays a more critical role than $A$ in encoding knowledge across clients.

\begin{figure}[h]
\begin{center}
\includegraphics[width=0.49\linewidth]{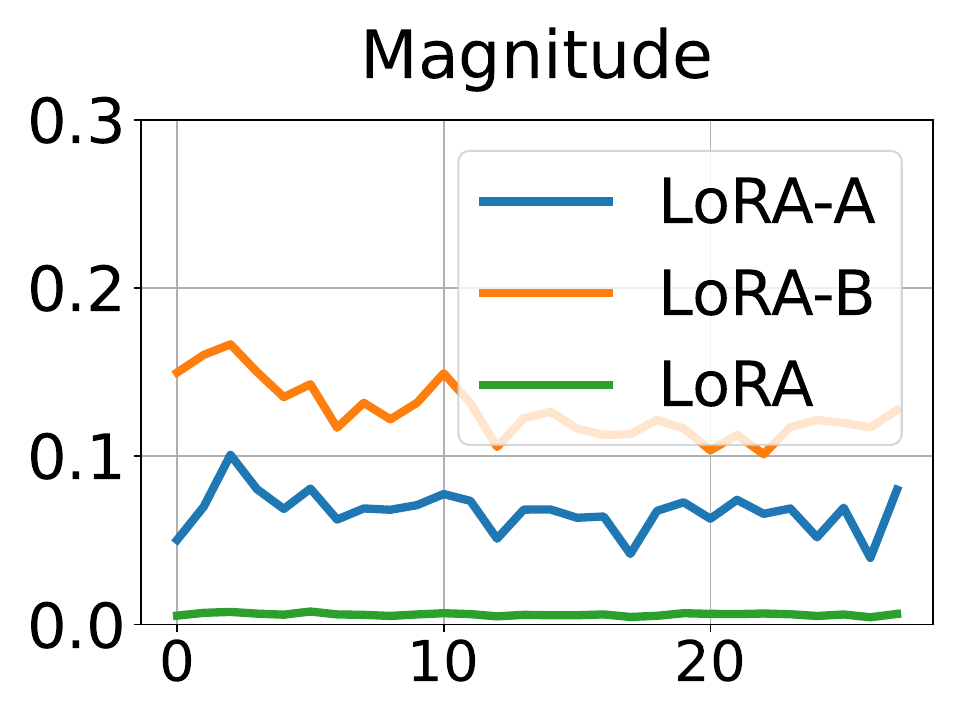}
\hspace{-1mm}
\includegraphics[width=0.49\linewidth]{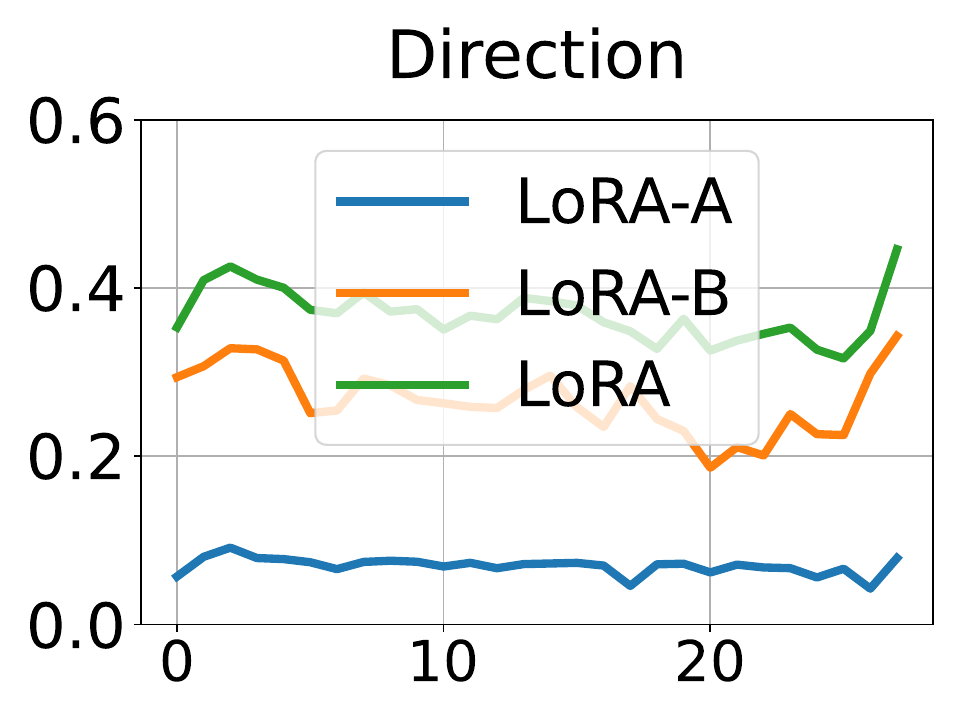}
\end{center}
\vspace{-0.5cm}
\caption{Comparison of LoRA modules using Qwen2-7B before and after fine-tuning. Left: magnitude change; Right: direction change.}
\label{fig:fig10}
\vspace{-0.4cm}
\end{figure}

To further explore whether the above observation depends on the model or dataset, we analyze the learning dynamics of LoRA the Qwen2-7B model using the bigscience/xP3 dataset \citep{muennighoff2023crosslingual}, which contains data from 46 languages and 16 NLP tasks. We sample 3,000 English examples and fine-tune the model using the same configuration. The results are shown in Figure~\ref{fig:fig10}. We observe the same pattern: the $B$ matrix plays a more dominant role than $A$ matrix during training.

\subsection{Practical Implementation}
For the results in Section~\ref{subsec:similarity}–\ref{subsec:dynamics}, we fine-tune LLaMA2-7B and Qwen2-7B for 3 epochs. The training uses a learning rate of 3e-4, batch size 32, and gradient accumulation step 2. We follow the alpaca\_short template \citep{alpaca} to construct the instruction data. LoRA is applied to the $q_\text{proj}$ modules with rank $r=8$.

For the analysis of lazy learning in multi-task fine-tuning (Section~\ref{subsec:sharing_A_vs_B}), we fine-tune the LLaMA3-8B model on the commonsense reasoning 15K dataset \citep{hu2023llm} for 3 epochs, using a learning rate of 3e-4, batch size 4, and gradient accumulation step 4. LoRA is applied to the $q_\text{proj}$ modules with rank $r=8$. The sharing-$A$ structure uses 3 $A$ matrices, and the sharing-$B$ structure uses 3 $B$ matrices.

For the analysis of knowledge transfer in federated fine-tuning (Section~\ref{subsec:sharing_A_vs_B}), we fine-tune the LLaMA2-7B model on the federated NLP dataset \citep{long2024dual} for 10 communication rounds, with 10 local epochs per client. The learning rate is 5e-4, the batch size is 32, and the gradient accumulation step is 2. In the homogeneous setting, all clients use rank 8. In the heterogeneous setting, the ranks are set to $\{64,64,32,32,16,16,8,8\}$. The methods are applied to $q_\text{proj}$ and $v_\text{proj}$ modules. All experiments are conducted on RTX A6000 GPU.

\section{Additional Experimental Details and Results}\label{appx:experiments}
\subsection{Benchmarks}\label{appx:benchmarks}

The intra-domain multi-task commonsense reasoning 170K benchmark contains questions from the following datasets: (1) ARC-Challenge
and ARC-Easy \citep{clark2018think}, which consist of grade-school–level multiple-choice science questions; (2) BoolQ \citep{clark2019boolq}, a yes/no question-answering dataset requiring non-factoid reasoning and entailment; (3) HellaSwag \citep{zellers2019hellaswag}, a dataset of commonsense natural language inference (NLI) questions that require identifying the most appropriate continuation of a narrative input; (4) OpenBookQA \citep{mihaylov2018can}, which contains questions requiring multi-step reasoning by combining provided scientific facts with external background knowledge; (5) PIQA \citep{bisk2020piqa}, a dataset of everyday commonsense reasoning questions about the physical world; (6) SIQA \citep{sap2019social}, which focuses on social and emotional commonsense reasoning in everyday human interactions; (7) WinoGrande \citep{sakaguchi2021winogrande}, a collection of fill-in-the-blank sentences designed to test pronoun resolution using commonsense.
 
The cross-domain multi-task NLP dataset contains 8 NLP tasks sampled from the FLAN dataset \citep{weifinetuned}. The tasks are: (1) Commonsense, a reasoning task that requires everyday knowledge to make judgments; (2) Entailment, an NLI task that determines the relationship between a premise and a hypothesis; (3) Open-domain QA, a question answering task that retrieves or generates answers from open sources; (4) Paraphrase, a classification task that recognizes whether a sentence pair is semantically equivalent; (5) Reading comprehension, a question answering task requires understanding the text content and answering the related questions; (6) Sentiment classification, a classification task that determines the whether the sentiment polarity is neutral, positive, or negative; (7) Summarization, an NLG task that produces a compact digest of a long passage while keeping the critical information; (8) Text formatting, an NLG task that corrects the punctuation in unformatted text. Each task has 300 examples for training and 200 examples for testing.

The federated NLP dataset also contains 8 NLP tasks sampled from the FLAN dataset \citep{weifinetuned}. The tasks are: (1) Coreference, a discourse understanding task that requires determining which entity a pronoun refers to; (2) Entailment, an NLI task that determines the relationship between a premise and a hypothesis; (3) Linguistic Acceptability, a classification task that detects whether a sentence is grammatical; (4) Paraphrase, a classification task that recognizes whether a sentence pair is semantically equivalent; (5) Question classification, a task for question understanding in question answering systems; (6) Structure-to-Text, a natural language generation (NLG) task that converts structured triples into natural language; (7) Text formatting, an NLG task that corrects the punctuation in unformatted text; (8) Word sense disambiguation, a classification task that determines whether the same word has the same meaning in two different sentences. Each task has 300 examples for training and 200 examples for testing, and we assign one task to each client. 

For the intra-domain multi-task fine-tuning, we also consider the math reasoning 10K benchmark \citep{hu2023llm}, which includes 4 datasets: (1) AQuA  \citep{ling2017program}, which contains multiple-choice algebra word problems, each accompanied by a natural language rationale explaining the step-by-step reasoning; (2) GSM8K \citep{cobbe2021training}, a high-quality collection of linguistically diverse grade-school–level math word problems designed to evaluate multi-step reasoning; (3) MAWPS \citep{koncel2016mawps}, a compilation of math word problems intended to support robust and scalable research on arithmetic reasoning, including AddSub (basic addition/subtraction), SingleOp (single-operator arithmetic), MultiArith (multi-step arithmetic), SingleEq (single-equation algebra); (4) SVAMP \citep{patel2021nlp}, which consists of simple one-unknown grade-school–level arithmetic word problems, designed to test robustness in arithmetic reasoning. 

The original split of the math reasoning 10K benchmark is not suitable for multi-task fine-tuning, since it does not include the full training data of the subsidiary tasks. In addition, \citet{hu2023llm} report data leakage issues in this benchmark. To address these concerns, we downloaded the original data of each single task, and checked every training example in the benchmark to determine whether it belongs to the training set of any individual task. This process allowed us to construct a training dataset for each task, making it possible to fine-tune on single tasks and obtain single-task baselines. For multi-task fine-tuning, we fine-tune directly on the benchmark and evaluate on each task. Since the single-task training splits are created by us, we report the corresponding results in the Appendix.

\subsection{Baselines}
For multi-task fine-tuning, we compare our ALoRA with the following methods: (1) the vanilla LoRA \citep{hulora}; (2) LoHa \citep{yeh2023navigating}, which employs Hadamard decompositions to the updates; (3) AdaLoRA \citep{zhangadaptive}, which adaptively prunes rank using SVD; (4) MoSLoRA \citep{wu2024mixture}, which introduces an additional matrix to fuse update subspaces; (5) HydraLoRA \citep{tian2024hydralora}, which adopts a sharing-$A$ multi-LoRA framework. 

For homogeneous federated fine-tuning, we compare our Fed-ALoRA with the following methods: (1) FedIT \citep{zhang2024towards}, where each client transmits the full LoRA parameters; (2) FedDPA \citep{long2024dual}, which employs both a global adapter and a local adapter for each client; (3) FedSA-LoRA \citep{guoselective}, which shares only the $A$ matrices. For the heterogeneous federated fine-tuning, we compare Fed-ALoRA with: (1) ZeroPadding, which pads all heterogeneous ranks to the maximum rank across clients, enabling FedIT to support heterogeneity; (2) FLoRA \citep{wang2024flora}, a stacking-based noise-free aggregation method; (3) FedSA-LoRA \citep{guoselective}, which does not natively support heterogeneity but is adapted here using the decomposition proposed in our method.

\subsection{Practical Implementation}

For intra-domain multi-task fine-tuning on commonsense reasoning and math reasoning, we fine-tune LLaMA3-8B and Qwen2-7B models for 3 epochs on the training data with a learning rate of 3e-4. The AdamW optimizer is used with $\beta_1=0.9$ and $\beta_2=0.999$. The batch size is 4, and the gradient accumulation step is 4. For HydraLoRA, we follow the original setup with one single $A$ matrix and 3 $B$ matrices with rank 8. For our ALoRA, we use 3 $A$ matrices and a single $B$ matrix with rank 8. To ensure a fair comparison, the other baselines are configured with a comparable parameter size: LoRA, LoHa, and MoSLoRA use rank 16, while others use rank 8. All methods are applied to the $q_\text{proj}$ and $o_\text{proj}$ modules. 

For cross-domain multi-task fine-tuning, we fine-tune LLaMA2-7B and Qwen2-7B models for 50 epochs on the training data with a learning rate of 3e-4. The AdamW optimizer is used with $\beta_1=0.9$ and $\beta_2=0.999$. The batch size is 4, and the gradient accumulation step is 4. For HydraLoRA, we follow the original setup with one single $A$ matrix and 3 $B$ matrices with rank 8. For our ALoRA, we use 3 $A$ matrices and a single $B$ matrix with rank 8. To ensure a fair comparison, the other baselines are configured with a comparable parameter size: LoRA, LoHa, and MoSLoRA use rank 16, while others use rank 8. All methods are applied to the $(q_\text{proj},v_\text{proj})$ modules. 

For federated fine-tuning, we fine-tune LLaMA2-7B and Qwen2-7B models for 10 communication rounds with 10 local epochs per client, using a learning rate of 5e-4. The AdamW optimizer is used with $\beta_1=0.9$ and $\beta_2=0.999$. The batch size is 32, and the gradient accumulation step is 2. In the homogeneous setting, all clients use rank 8. In the heterogeneous setting, the ranks are set to $\{64,64,32,32,16,16,8,8\}$, and $d_m$ is chosen to 16. All methods are applied to $(q_\text{proj},v_\text{proj})$ modules for LLaMA2-7B and to $(q_\text{proj},o_\text{proj})$ modules for Qwen2-7B model. All experiments are conducted on RTX A6000 GPU.

\subsection{Full Results on Sharing-$A$ and Sharing-$B$}
\label{appx:extra_exps_sharing_A_B}

To assess which parameters should be shared to improve cross-client knowledge transder, we compare the performance of each client on all tasks between FedSA and our Fed-ALoRA in Section~\ref{subsec:sharing_A_vs_B}. The full results are presented in Table~\ref{tab:full_cross_client_transfer}.

\begin{table*}[ht]
\begin{center}
\begin{adjustbox}{width=0.9\linewidth}
\begin{tabular}{ccccccccccc}
\toprule
Setting & Method & Client 1 & Client 2 & Client 3 & Client 4 & Client 5 & Client 6 & Client 7 & Client 8 & Avg. \\
\midrule
\multirow{2}{*}{Homogeneous}
& Sharing $A$ & 50.02 & 49.78 & 59.92 & 42.66 & 54.05 & 25.21 & 23.32 & 49.43 & 44.30 \\
& Sharing $B$  & 67.43 & 67.83 & 69.85 & 69.26 & 69.50 & 60.94 & 54.86 & 70.90 & 66.32 \\
\midrule
\multirow{2}{*}{Heterogeneous}
& Sharing $A$ & 43.01 & 41.58 & 50.09 & 38.34 & 53.53 & 24.82 & 24.28 & 50.41 & 40.76 \\
& Sharing $B$  & 48.38 & 54.80 & 58.30 & 46.98 & 62.15 & 35.09 & 31.80 & 64.89 & 50.30 \\
\bottomrule
\end{tabular}
\end{adjustbox}
\end{center}
\vspace{-0.2cm}
\caption{Comparing sharing $A$ versus $B$ in federated fine-tuning. Sharing $B$ consistently outperforms sharing $A$ in both the homogeneous and heterogeneous settings.}
\label{tab:full_cross_client_transfer}
\end{table*}

\subsection{Full Results on Commonsense Reasoning}
\label{appx:extra_exps_commonsense}

We provide the full experiment results on the commonsense reasoning benchmark, including the per-task performance, mean and standard deviations of 3 independent runs. The details are shown in Table~\ref{tab:full_commonsense}.

\begin{table*}[t]
\begin{center}
\begin{adjustbox}{width=\linewidth}
\begin{tabular}{c|cccccccc|cc}
\toprule
Method & ARC-C & ARC-E & BoolQ & HellaS. & OBQA & PIQA & SIQA & WinoG. & Avg. & $\Delta m\%(\downarrow)$ \\
\midrule
\multicolumn{11}{c}{\textit{LLaMA3-8B}} \\
\midrule
ST Baseline & 77.76 $_{\pm0.97}$ & 90.81 $_{\pm0.23}$ & 73.39 $_{\pm1.05}$ & 95.40 $_{\pm0.08}$ & 86.60 $_{\pm0.72}$ & 89.25 $_{\pm0.47}$ & 80.69 $_{\pm0.48}$ & 85.35 $_{\pm0.64}$ & 84.90 $_{\pm 0.21}$ & \\
LoRA & 76.66 $_{\pm1.02}$ & 88.72 $_{\pm0.83}$ & 72.73 $_{\pm1.24}$ & 94.50 $_{\pm0.41}$ & 84.10 $_{\pm0.14}$ & 87.59 $_{\pm0.15}$ & 79.42 $_{\pm0.15}$ & 85.40 $_{\pm1.00}$ & 83.64 $_{\pm0.27}$ & 1.48 \\
LoHa & 77.13 $_{\pm0.13}$ & 88.91 $_{\pm0.09}$ & 72.89 $_{\pm0.19}$ & 94.05 $_{\pm0.23}$ & 85.40 $_{\pm0.56}$ & 87.84 $_{\pm0.34}$ & 78.92 $_{\pm0.44}$ & 84.73 $_{\pm0.17}$ & 83.73 $_{\pm0.11}$ & 1.36 \\
AdaLoRA & 77.72 $_{\pm0.72}$ & 89.72 $_{\pm0.96}$ & 73.32 $_{\pm0.19}$ & 93.98 $_{\pm0.73}$ & 85.20 $_{\pm1.13}$ & 87.90 $_{\pm0.35}$ & 79.45 $_{\pm0.18}$ & 83.78 $_{\pm0.28}$ & 83.88 $_{\pm0.32}$ & 1.17 \\
MoSLoRA & 76.87 $_{\pm0.12}$ & 89.22 $_{\pm0.15}$ & 73.50 $_{\pm0.86}$ & 95.02 $_{\pm0.13}$ & 85.27 $_{\pm2.58}$ & 87.99 $_{\pm0.59}$ & 80.89 $_{\pm0.12}$ & 85.13 $_{\pm0.55}$ & 84.23 $_{\pm0.39}$ & 0.76 \\
HydraLoRA & 78.58 $_{\pm0.12}$ & 89.94 $_{\pm0.18}$ & 75.02 $_{\pm0.18}$ & 95.21 $_{\pm0.11}$ & 84.90 $_{\pm1.55}$ & 88.03 $_{\pm0.15}$ & 79.99 $_{\pm0.72}$ & 84.92 $_{\pm0.34}$ & 84.57 $_{\pm0.17}$ & 0.32 \\
\rowcolor{oursblue}
ALoRA (ours) & 79.40 $_{\pm0.41}$ & 89.69 $_{\pm0.71}$ & 74.38 $_{\pm0.10}$ & 94.85 $_{\pm0.23}$ & 86.10 $_{\pm0.14}$ & 88.24 $_{\pm0.49}$ & 80.17 $_{\pm0.32}$ & 85.68 $_{\pm0.27}$ & \textbf{84.81} $_{\pm0.03}$ & \textbf{0.04}\\
\midrule
\multicolumn{11}{c}{\textit{Qwen2-7B}} \\
\midrule
ST Baseline & 86.40 $_{\pm0.38}$ & 94.01 $_{\pm0.14}$ & 74.04 $_{\pm1.01}$ & 95.54 $_{\pm0.25}$ & 91.33 $_{\pm0.83}$ & 90.91 $_{\pm0.33}$ & 81.94 $_{\pm0.15}$ & 83.97 $_{\pm0.85}$ & 87.27 $_{\pm0.24}$ & \\
LoRA & 83.30 $_{\pm0.55}$ & 92.95 $_{\pm0.12}$ & 74.37 $_{\pm0.40}$ & 93.68 $_{\pm1.20}$ & 88.46 $_{\pm1.22}$ & 89.13 $_{\pm0.49}$ & 80.48 $_{\pm0.29}$ & 85.34 $_{\pm0.43}$ & 85.96 $_{\pm0.31}$ & 1.43 \\
LoHa & 85.22 $_{\pm0.72}$ & 94.03 $_{\pm0.28}$ & 73.58 $_{\pm0.11}$ & 94.38 $_{\pm0.52}$ & 88.53 $_{\pm0.61}$ & 89.85 $_{\pm0.37}$ & 80.31 $_{\pm0.22}$ & 84.76 $_{\pm0.85}$ & 86.34 $_{\pm0.33}$ & 1.06 \\
AdaLoRA & 85.39 $_{\pm0.05}$ & 93.86 $_{\pm0.06}$ & 72.61 $_{\pm0.35}$ & 94.09 $_{\pm0.30}$ & 89.60 $_{\pm0.20}$ & 89.13 $_{\pm0.03}$ & 80.95 $_{\pm0.11}$ & 83.50 $_{\pm0.83}$ & 86.14 $_{\pm0.17}$ & 1.30 \\
MoSLoRA & 84.80 $_{\pm0.64}$ & 93.18 $_{\pm0.19}$ & 73.33 $_{\pm1.13}$ & 94.02 $_{\pm0.71}$ & 89.80 $_{\pm0.53}$ & 89.66 $_{\pm0.34}$ & 80.58 $_{\pm0.67}$ & 85.05 $_{\pm0.81}$ & 86.30 $_{\pm0.16}$ & 1.09 \\
HydraLoRA & 84.60 $_{\pm0.30}$ & 93.20 $_{\pm0.03}$ & 73.37 $_{\pm1.42}$ & 94.42 $_{\pm0.37}$ & 88.30 $_{\pm0.42}$ & 89.58 $_{\pm0.73}$ & 80.97 $_{\pm0.06}$ & 84.33 $_{\pm0.95}$ & 86.09 $_{\pm0.42}$ & 1.32 \\
\rowcolor{oursblue}
ALoRA (ours) & 85.28 $_{\pm0.42}$ & 93.69 $_{\pm0.29}$ & 73.41 $_{\pm0.15}$ & 94.76 $_{\pm0.36}$ & 90.10 $_{\pm0.42}$ & 89.07 $_{\pm0.07}$ & 80.94 $_{\pm0.47}$ & 84.50 $_{\pm0.50}$ & \textbf{86.47} $_{\pm0.07}$ & \textbf{0.91} \\
\bottomrule
\end{tabular}
\end{adjustbox}
\end{center}
\vspace{-0.2cm}
\caption{Results on intra-domain multi-task commonsense reasoning benchmark. $\Delta m\%$ measures performance balance across tasks. $\downarrow$ denotes that lower values are better.  All methods use the same number of adapter parameters. We run each experiment 3 times and report the average with the error bar.}
\label{tab:full_commonsense}
\end{table*}

\subsection{Full Results on Multi-Task NLP Datasets}
\label{appx:extra_exps_mtl_nlp}

We provide the full experiment results on the cross-domain multi-task NLP datasets, including the per-task performance. The details are shown in Table~\ref{tab:full_mtl_nlp}.

\begin{table*}[t]
\begin{center}
\begin{adjustbox}{width=0.8\linewidth}
\begin{tabular}{c|cccccccc|cc}
\toprule
Method & CSR & Ent & ODQA & Para
       & RC & Sent & Sum & TFmt 
       & Avg. & $\Delta m\%$ \\

\midrule
\multicolumn{11}{c}{\textit{LLaMA2-7B}} \\
\midrule
ST Baseline & 45.15 & 65.00 & 75.19 & 55.00 & 78.00 & 71.75 & 28.17 & 88.6 & 63.36 & \\
LoRA & 53.19 & 63.00 & 84.31 & 51.00 & 50.50 & 69.50 & 32.53 & 89.36 & 61.67 & 0.31 \\
LoHa & 49.94 & 60.94 & 79.78 & 67.70 & 69.57 & 73.50 & 33.33 & 90.85 & 65.70 & -5.76 \\
AdaLoRA & 51.94 & 56.94 & 78.57 & 65.22 & 66.61 & 59.50 & 31.95 & 84.93 & 61.96 & -0.41 \\
MoSLoRA & 50.70 & 60.50 & 81.11 & 71.50 & 70.00 & 75.00 & 32.64 & 87.95 & 66.18 & -6.58 \\
HydraLoRA & 44.51 & 67.50 & 75.83 & 74.50 & 76.50 & 71.50 & 32.14 & 89.10 & 66.45 & -6.39 \\
\rowcolor{oursblue}

ALoRA  & 48.21 & 62.50 & 80.35 & 78.50 & 68.50 & 75.00 & 33.79 & 90.20 & \textbf{67.13} & \textbf{-8.33} \\
\midrule
\multicolumn{11}{c}{\textit{Qwen2-7B}} \\
\midrule
ST Baseline & 76.30 & 90.00 & 90.95 & 87.5 & 77.50 & 71.00 & 25.46 & 90.59 & 76.16 & \\
LoRA & 87.20 & 86.00 & 94.41 & 86.00 & 80.50 & 76.50 & 32.35 & 93.87 & 79.60 & -6.78 \\
LoHa & 84.52 & 89.00 & 93.65 & 82.00 & 76.50 & 76.00 & 32.54 & 92.62 & 78.35 & -5.27 \\
AdaLoRA & 81.11 & 87.00 & 93.92 & 82.50 & 74.55 & 76.00 & 32.57 & 93.26 & 77.61 & -4.33 \\
MoSLoRA & 84.75 & 89.50 & 94.63 & 86.00 & 80.06 & 72.50 & 33.26 & 94.02 & 79.34 & -6.59 \\
HydraLoRA & 88.16 & 87.00 & 93.84 & 88.50 & 80.23 & 76.50 & 31.95 & 94.04 & 80.03 & -7.14 \\
\rowcolor{oursblue}

ALoRA & 84.99 & 89.50 & 94.79 & 85.50 & 81.50 & 80.00 & 32.77 & 94.65 & \textbf{80.46} & \textbf{-7.98} \\
\bottomrule
\end{tabular}
\end{adjustbox}
\end{center}
\vspace{-0.2cm}
\caption{Results on cross-domain multi-task NLP datasets. $\Delta m\%$ measures performance.}
\label{tab:full_mtl_nlp}
\end{table*}

\subsection{Full Results on Homogeneous Federated Learning}
\label{appx:extra_exps_homo}

We provide the full experiment results on the homogeneous federated learning, including the per-client performance. The details are shown in Table~\ref{tab:full_homo}.

\begin{table*}[t]
\begin{center}
\begin{adjustbox}{width=0.9\linewidth}
\begin{tabular}{c|cccccccc|ccc}
\toprule
Method & Coref & Ent & LAcc & Para
       & QCls & S2T & TFmt & WSD 
       & Avg. & $\Delta m\%$ & Params. \\
\midrule
\multicolumn{12}{c}{\textit{LLaMA2-7B}} \\
\midrule
ST Baseline & 73.00 & 84.00 & 79.00 & 78.00 & 94.00 & 72.21 & 96.64 & 60.50 & 79.67 & & \\
FedIT & 86.24 & 86.50 & 78.00 & 81.00 & 94.50 & 72.06 & 96.51 & 65.00 & 82.47 & -3.92 & 8.39 \\
FedDPA & 88.51 & 85.50 & 73.50 & 77.50 & 95.50 & 73.76 & 96.40 & 65.00 & 81.96 & -3.30 & 16.78 \\
FedSA-LoRA & 81.77 & 86.00 & 78.00 & 75.00 & 93.50 & 73.34 & 96.55 & 65.00 & 81.15 & -2.21 & \textbf{4.19} \\
\rowcolor{oursblue}

Fed-ALoRA  & 85.74 & 87.00 & 73.50 & 79.00 & 94.00 & 73.10 & 96.24 & 71.50 & \textbf{82.51} & \textbf{-4.29} & \textbf{4.19} \\
\midrule
\multicolumn{12}{c}{\textit{Qwen2-7B}} \\
\midrule
ST Baseline & 84.08 & 91.50 & 80.00 & 78.00 & 91.00 & 72.83 & 97.11 & 67.50 & 82.75 & & \\
FedIT & 82.96 & 92.00 & 79.00 & 82.00 & 91.00 & 71.60 & 96.86 & 67.00 & 82.80 & -0.05 & 6.42 \\
FedDPA & 83.41 & 93.50 & 80.00 & 79.50 & 88.00 & 73.83 & 97.58 & 69.50 & 83.17 & -0.60 & 12.85 \\
FedSA-LoRA & 85.58 & 91.00 & 82.50 & 79.50 & 93.00 & 72.76 & 97.18 & 68.00 & 83.69 & -1.15 & \textbf{3.21} \\
\rowcolor{oursblue}

Fed-ALoRA  & 84.02 & 91.50 & 80.00 & 83.50 & 93.00 & 73.64 & 97.27 & 71.50 & \textbf{84.30} & \textbf{-2.05} & \textbf{3.21} \\
\bottomrule
\end{tabular}
\end{adjustbox}
\end{center}
\vspace{-0.2cm}
\caption{Results for the {\bf homogeneou}s federated setting. Params. denotes the average number of parameters (in millions) transmitted per client in each round. ALoRA achieves the most balanced performance while reducing communication cost by 50\% compared to full LoRA aggregation FedIT.}
\label{tab:full_homo}
\end{table*}

\subsection{Full Results on Heterogeneous Federated Learning}
\label{appx:extra_exps_hetero}

We provide the full experiment results on the heterogeneous federated learning, including per-client performance. We implement the Sharing-$A$ method based on our decomposition strategy for dealing with varying clients. The details are shown in Table~\ref{tab:full_hetero}.

\begin{table*}[t]
\begin{center}
\begin{adjustbox}{width=0.9\linewidth}
\begin{tabular}{c|cccccccc|ccc}
\toprule
Method & Coref & Ent & LAcc & Para
       & QCls & S2T & TFmt & WSD 
       & Avg. & $\Delta m\%$ & Params.\\
\midrule
\multicolumn{12}{c}{\textit{LLaMA2-7B}} \\
\midrule
ST Baseline & 81.62 & 88.00 & 81.00 & 79.50 & 94.50 & 72.07 & 96.64 & 60.50 & 81.73 & &\\
ZeroPadding & 86.95 & 87.00 & 77.50 & 79.50 & 94.00 & 72.87 & 96.46 & 64.00 & 82.29 & -0.91 & 49.28 \\
FLoRA & 82.03 & 87.50 & 75.50 & 74.00 & 95.50 & 70.99 & 96.07 & 62.00 & 80.45 & 1.54 & 141.56 \\
Sharing-$A$ & 80.26 & 83.50 & 76.50 & 76.00 & 93.00 & 73.49 & 96.61 & 54.00 & 79.17 & 3.39 & \textbf{12.12} \\
\rowcolor{oursblue}

Fed-ALoRA & 88.27 & 89.50 & 79.50 & 77.00 & 94.00 & 72.35 & 96.37 & 63.00 & \textbf{82.50} & \textbf{-1.07} & \textbf{12.12} \\
\midrule
\multicolumn{12}{c}{\textit{Qwen2-7B}} \\
\midrule
ST Baseline & 84.25 & 90.50 & 81.00 & 82.00 & 92.00 & 73.05 & 97.11 & 67.50 & 83.43 & &\\
ZeroPadding & 84.07 & 92.50 & 80.50 & 79.01 & 90.00 & 73.01 & 97.38 & 70.05 & 83.32 & 0.06 & 37.73 \\
FLoRA & 79.98 & 92.00 & 80.00 & 77.00 & 88.00 & 71.04 & 97.19 & 68.00 & 81.65 & 2.13 & 108.38 \\
Sharing-$A$ & 85.16 & 90.50 & 79.50 & 78.00 & 90.50 & 74.32 & 97.43 & 69.00 & 83.05 & 0.37 & \textbf{9.23} \\
\rowcolor{oursblue}

Fed-ALoRA & 85.50 & 91.50 & 82.50 & 80.02 & 90.00 & 74.49 & 97.51 & 71.50 & \textbf{84.13} & \textbf{-1.02} & \textbf{9.23} \\
\bottomrule
\end{tabular}
\end{adjustbox}
\end{center}
\vspace{-0.2cm}
\caption{Results for the {\bf heterogeneous} setting. ALoRA achieves the most balanced performance while reducing communication cost by 75\% compared to full LoRA aggregation ZeroPadding. The original FedSA-LoRA does not support heterogeneity; Sharing-$A$ denotes implementation with our decomposition strategy.}
\label{tab:full_hetero}
\end{table*}

\subsection{Full Results on the Intermediate Rank}
We also provide the full results of the study of different intermediate ranks using LLaMA2-7B model in Section~\ref{exp:depth_analysis}, which are shown in Table~\ref{tab:extra_exps_intermediate_rank}.

\begin{table*}[ht]
\begin{center}
\begin{adjustbox}{width=0.8\linewidth}
\begin{tabular}{c|cccccccc|cc}
\toprule
Fed-ALoRA & Coref & Ent & LAcc & Para
       & QCls & S2T & TFmt & WSD 
       & Avg. & $\Delta m\%(\downarrow)$\\
\midrule
$d_m=8$ & 90.75 & 84.50 & 79.50 & 73.50 & 95.00 & 73.05 & 96.09 & 63.00 & 81.92 & -0.41 \\
$d_m=16$ & 88.27 & 89.50 & 79.50 & 77.00 & 94.00 & 72.35 & 96.37 & 63.00 & \textbf{82.50} & \textbf{-1.07} \\
$d_m=32$ & 88.63 & 86.50 & 78.00 & 79.00 & 94.00 & 72.27 & 96.60 & 63.00 & 82.25 & -0.80 \\
$d_m=64$ & 85.70 & 89.00 & 81.00 & 75.50 & 94.00 & 72.31 & 96.40 & 65.00 & 82.37 & -1.01 \\
\bottomrule
\end{tabular}
\end{adjustbox}
\end{center}
\caption{Results of LLaMA2-7B on different intermediate ranks in the heterogeneous setting.}
\label{tab:extra_exps_intermediate_rank}
\end{table*}

\subsection{Comparison with Additional Multi-Task Fine-Tuning Methods}
\label{appx:comparision_extra_mtft}

We compare ALoRA with existing multi-task fine-tuning methods that explicitly utilizing task-specific information. Since MASA has not publicly released its source code, we are unable to include it in our comparison. The results shown in Table~\ref{tab:comparison_extra_mtft} further demonstrate the effectiveness of our method.

\begin{table*}[t]
\begin{center}
\begin{adjustbox}{width=0.8\linewidth}
\begin{tabular}{c|cccccccc|cc}
\toprule
Method & ARC-C & ARC-E & BoolQ & HellaS. & OBQA & PIQA & SIQA & WinoG. & Avg. & $\Delta m\%(\downarrow)$ \\
\midrule
ST Baseline & 77.76 & 90.81 & 73.39 & 95.40 & 86.60 & 89.25 & 80.69 & 85.35 & 84.90 & \\
\midrule
LoRAMoE & 76.53 & 88.01 & 73.34 & 93.97 & 83.80 & 86.24 & 80.31 & 83.57 & 83.22 & 1.92 \\
MTL-LoRA & 79.69 & 89.50 & 74.47 & 94.88 & 85.81 & 87.95 & 80.67 & 85.72 & 84.84 & -0.001 \\
CoLA & 79.30 & 89.09 & 74.17 & 94.78 & 85.19 & 87.48 & 79.61 & 85.06 & 84.34 & 0.60 \\
\midrule
ALoRA (ours) & 79.40  & 89.69  & 74.38  & 94.85  & 86.10  & 88.24  & 80.17  & 85.68  & 84.81  & 0.04\\

\bottomrule
\end{tabular}
\end{adjustbox}
\end{center}
\vspace{-0.2cm}
\caption{Comparison on commonsense benchmarks between ALoRA and multi-task fine-tuning methods that explicitly use task information. Experiments are conducted on LLaMA3-8B.}
\label{tab:comparison_extra_mtft}
\end{table*}

\subsection{Comparison Between HydraLoRA and ALoRA at Larger Model Scales}
\label{appx:comparison_large_scale}

To further validate the benefits of sharing $B$ at a larger model scale, we conduct experiments on commonsense benchmark using Qwen2.5-14B. We adopt a learning rate of 3e-5 and apply adaptation to the $q_\text{proj}$ and $o_\text{proj}$ modules. The results are shown in Table~\ref{tab:comparison_extra_scale}. They show that ALoRA consistently outperforms HydraLoRA.

\begin{table*}[t]
\begin{center}
\begin{adjustbox}{width=0.8\linewidth}
\begin{tabular}{c|cccccccc|c}
\toprule
Method & ARC-C & ARC-E & BoolQ & HellaS. & OBQA & PIQA & SIQA & WinoG. & Avg. \\
\midrule
HydraLoRA & 92.19 & 97.43 & 74.89 & 95.50 & 91.41 & 92.76 & 82.62 & 89.24 & 89.50 \\
ALoRA (ours) & 92.24  & 97.58 & 74.86 & 95.74 & 91.60 & 92.82 & 82.75 & 89.15 & 89.59 \\

\bottomrule
\end{tabular}
\end{adjustbox}
\end{center}
\vspace{-0.2cm}
\caption{Comparison between ALoRA and HydraLoRA on commonsense benchmarks at a larger model scale.}
\label{tab:comparison_extra_scale}
\end{table*}

\subsection{Additional Results on Math Reasoning}
The results of LLaMA2-7B model on the math reasoning benchmark are presented in Table~\ref{tab:extra_exps_math_reasoning}. LoRA, MoSLoRA, and our ALoRA outperform the single-task baseline on all tasks, but ALoRA achieves the most balanced performance, showing that it enables more effective knowledge transfer than the baselines. 

\begin{table*}[t]
\begin{center}
\begin{adjustbox}{width=0.8\linewidth}
\begin{tabular}{c|ccccc|cc}
\toprule
Method & AQuA & GSM8K & SVAMP & MAWPS & SingleEq & Avg. & $\Delta m\%(\downarrow)$ \\
\midrule
ST Baseline & 28.34 & 63.68 & 71.10 & 86.13 & 90.75 & 68.00 & \\
\midrule
LoRA & 30.31 & 66.26 & 75.10 & 90.34 & 94.88 & 71.38 & -5.21 \\
LoHa & 27.56 & 63.84 & 76.70 & 90.76 & 94.88 & 70.75 & -3.06 \\
AdaLoRA & 25.20 & 59.44 & 72.60 & 86.55 & 91.93 & 67.14 & 2.77 \\
MoSLoRA & 28.74 & 67.40 & 77.10 & 89.08 & 94.06 & 71.28 & -4.54 \\
\midrule
HydraLoRA & 27.17 & 68.76 & 75.60 & 90.34 & 93.31 & 71.04 & -3.58 \\
\rowcolor{oursblue}
ALoRA & 29.53 & 67.17 & 77.40 & 89.50 & 94.49 & \textbf{71.62} & \textbf{-5.31} \\
\bottomrule
\end{tabular}
\end{adjustbox}
\end{center}
\vspace{-0.2cm}
\caption{Results of LLaMA2-7B on math reasoning. $\Delta m\%$ measures performance balance across tasks. ALoRA achieves the most balanced results.}
\label{tab:extra_exps_math_reasoning}
\end{table*}

\section{The Use of Large Language Models}
In preparing this manuscript, large language models (LLMs) were used only to assist with language
polishing and stylistic refinement. All technical content, formulations, experimental designs, and
conceptual contributions were developed by the authors. Importantly, LLMs were not used for
ideation and methodology development.

\end{document}